\documentclass[journal]{IEEEtran} 

\usepackage{cite}

\ifCLASSINFOpdf
  \usepackage[pdftex]{graphicx}
\else
\fi
\usepackage{amsmath}
\usepackage{array}

\ifCLASSOPTIONcompsoc
  \usepackage[caption=false,font=normalsize,labelfont=sf,textfont=sf]{subfig}
\else
  \usepackage[caption=false,font=footnotesize]{subfig}
\fi
\usepackage{url}

\newcommand{\Tref}[1]{Table~\ref{#1}}
\newcommand{\Eref}[1]{Eq.~(\ref{#1})}
\newcommand{\Fref}[1]{Fig.~\ref{#1}}
\newcommand{\Aref}[1]{Alg.~\ref{#1}}
\newcommand{\Sref}[1]{Section~\ref{#1}}
\newcommand{\Appref}[1]{Appendix~\ref{#1}}
\DeclareMathOperator*{\argmin}{arg\,min}

\graphicspath{{./img/}}

\newcommand{\bvec}[1]{\mathbf{#1}}
\usepackage{amssymb}

\usepackage[ruled, vlined, linesnumbered]{algorithm2e} 
\usepackage[usenames, dvipsnames]{color} 
\newcommand{\codecommentcolor}{OliveGreen} 
\newcommand{\codecomment}[1]{\textcolor{\codecommentcolor}{\texttt{#1}}}

\usepackage{color}

\usepackage{booktabs} 
\usepackage{multirow} 

\renewcommand{\baselinestretch}{0.9685}

\begin{document}
%
\title{PQTable: Non-exhaustive Fast Search for Product-quantized Codes using Hash Tables}
%
%
%

\author{Yusuke~Matsui,~\IEEEmembership{Member,~IEEE,}
        Toshihiko~Yamasaki,~\IEEEmembership{Member,~IEEE,}
        and~Kiyoharu~Aizawa,~\IEEEmembership{Fellow,~IEEE}
\thanks{Y. Matsui is with National Institute of Informatics, Tokyo, Japan.
	e-mail: matsui@nii.ac.jp}
\thanks{T. Yamasaki and K. Aizawa are with the Department of Information and Communication Engineering, the University of Tokyo, Tokyo, Japan.
	e-mail: \{yamasaki, aizawa\}@hal.t.u-tokyo.ac.jp
	}}

%
%

\markboth{Journal of \LaTeX\ Class Files,~Vol.~14, No.~8, August~2015}%
{Shell \MakeLowercase{\textit{et al.}}: Bare Demo of IEEEtran.cls for IEEE Journals}
%



\maketitle

\begin{abstract}
In this paper, we propose a product quantization table (PQTable); a fast search method for product-quantized codes via hash-tables.
An identifier of each database vector is associated with the slot of a hash table by using its PQ-code as a key.
For querying, an input vector is PQ-encoded and hashed, and the items associated with that code are then retrieved.
The proposed PQTable produces the same results as a linear PQ scan, and is $10^2$ to $10^5$ times faster.
Although state-of-the-art performance can be achieved by previous inverted-indexing-based approaches,
such methods require manually-designed parameter setting and significant training; our PQTable is free of these limitations, and therefore offers a practical and effective solution for real-world problems.
Specifically, when the vectors are highly compressed, our PQTable achieves one of the fastest search performances on a single CPU to date with significantly efficient memory usage (0.059 ms per query over $10^9$ data points with just 5.5 GB memory consumption).
Finally, we show that our proposed PQTable can naturally handle the codes of an optimized product quantization (OPQTable).

\end{abstract}


%
\IEEEpeerreviewmaketitle

\section{Introduction}
\begin{figure*}
	\begin{center}
		\subfloat[Linear ADC scan]{\includegraphics[height=0.13\linewidth]{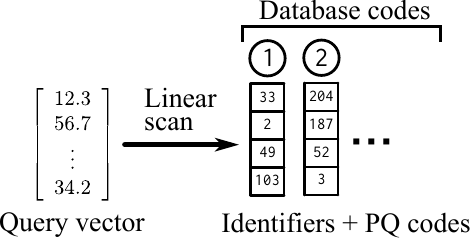}
			\label{fig:teaser_fig_adc}}
		\qquad
		\subfloat[Short-code-based inverted indexing system]{\includegraphics[height=0.1525\linewidth]{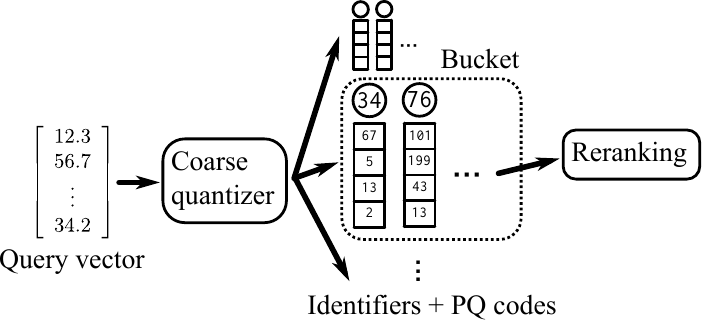}
			\label{fig:teaser_fig_invindex}}
		\qquad
		\subfloat[Proposed PQTable]{\includegraphics[height=0.145\linewidth]{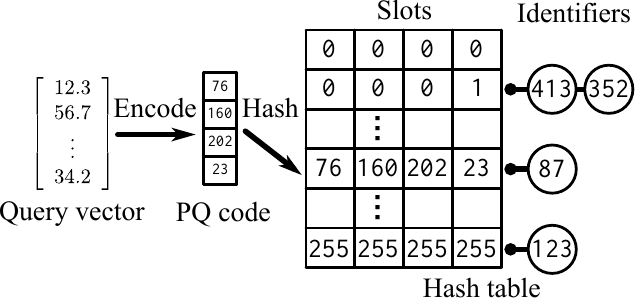}
			\label{fig:teaser_fig}}
	\end{center}
	\caption{Data structures of ANN systems: linear ADC scan, short-code-based inverted indexing systems, and PQTable.}
	\label{fig:teaser}
\end{figure*}

\IEEEPARstart{W}{ith}
the explosive growth of multimedia data, compressing high-dimensional vectors and performing approximate nearest neighbor (ANN) searches
in the compressed domain is becoming a fundamental problem when handling large databases.
\textit{Product quantization (PQ)}~\cite{tpami_jegou2011},
and its extensions~\cite{cvpr_norouzi2013, tpami_ge2014, cvpr_babenko2014, icml_zhang2014, tkde_wang2015, cvpr_heo2014, cvpr_babenko2015, cvpr_zhang2015, eccv_jain2016, tkde_ozan2016, tmm_ning2017},
are popular and successful methods for quantizing a vector into a short code.
PQ has three attractive properties:
(1) PQ can compress an input vector into an extremely short code (e.g., 32 bit); 
(2) the approximate distance between a raw vector and a compressed PQ code can be computed efficiently (the so-called~\textit{asymmetric distance computation (ADC)}~\cite{tpami_jegou2011}), which is a good estimate of the original Euclidean distance;
and (3) the data structure and coding algorithms are surprisingly simple.
Typically, database vectors are quantized into short codes in advance.
When given a query vector, similar vectors can be found 
from the database codes via a linear comparison 
using ADC (see \Fref{fig:teaser_fig_adc}).

Although 
linear ADC scanning is simple and easy to execute,
it is efficient only for small datasets as the search is exhaustive
(the computational cost is at least $O(N)$ for $N$ PQ codes).
To handle large (e.g., $N \sim 10^9$) databases,
\textit{short-code-based inverted indexing systems}~\cite{cvpr_babenko2012, iccv_xia2013, tpami_ge2014, cvpr_kalantidis2014, corr_babenko2014, tpami_babenko2015, cvpr_heo2016}
have been proposed, which are currently the state-of-the-art ANN methods (see \Fref{fig:teaser_fig_invindex}).
These systems operate in two stages:
(1) coarse quantization and (2) reranking via short codes.
In the data indexing phase, each database vector is first assigned to a cell using a coarse quantizer
(e.g., k-means~\cite{tpami_jegou2011}, multiple k-means~\cite{iccv_xia2013}, or Cartesian products~\cite{cvpr_babenko2012, iccv_iwamura2013}).
Next, the residual difference between the database vector and the coarse centroid is compressed
to a short code using PQ~\cite{tpami_jegou2011} or its extensions~\cite{cvpr_norouzi2013, tpami_ge2014}.
Finally, the code is stored as a posting list in the cell.
In the retrieval phase, a query vector is assigned to the nearest cells by the coarse quantizer, and associated items in corresponding posting lists are traversed, with the nearest one being reranked via ADC.
These systems are 
fast, accurate, and memory efficient,
as they can hold 
$10^9$ data points in memory and can conduct a retrieval in milliseconds.

\begin{figure}[t]
	\begin{minipage}{0.24\textwidth}
		\includegraphics[width=1.0\linewidth]{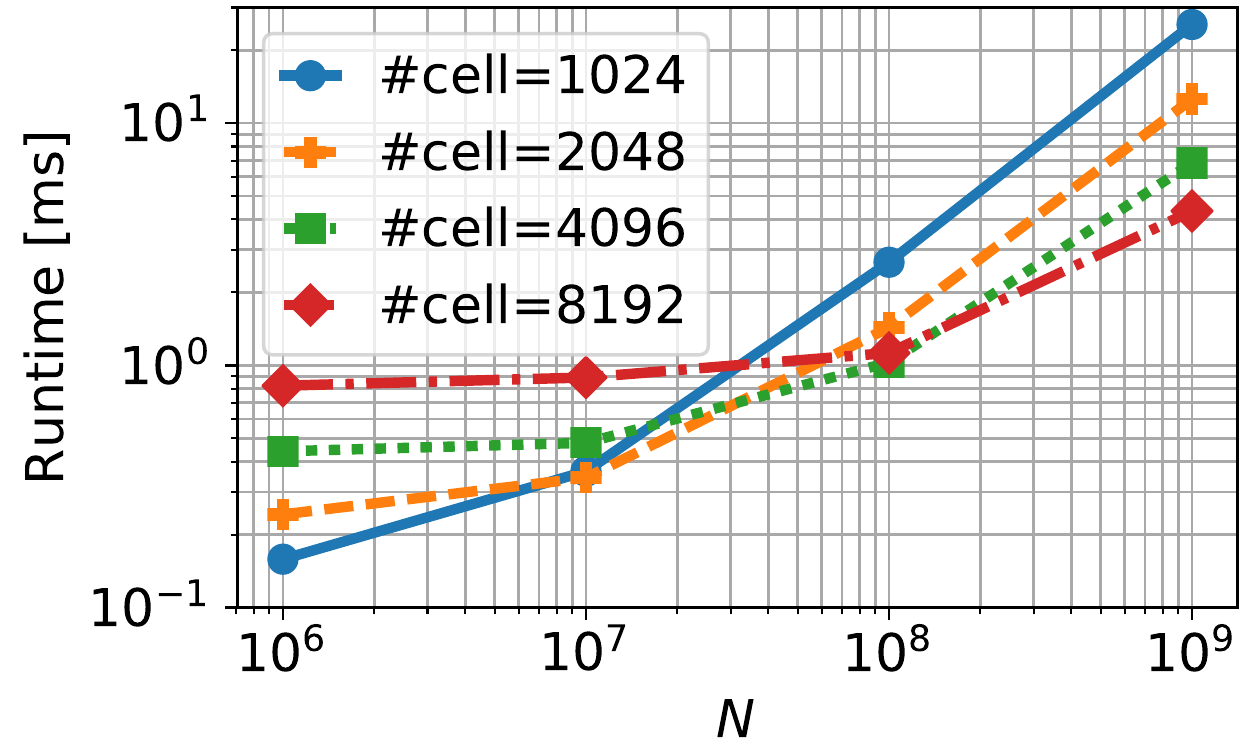}
	\end{minipage}~~~
	\begin{minipage}{0.24\textwidth}
		\scalebox{0.8}{	
			\begin{tabular}{@{}llll@{}} \toprule
				& \multicolumn{3}{c}{$w$} \\ \cmidrule(l){2-4}
				& 1 & 8 & 64 \\ \midrule
				Recall@1 & 0.121 & 0.146 & 0.147 \\
				Runtime & 2.7 & 20.5 & 177 \\ \bottomrule
			\end{tabular}
		}
	\end{minipage}
	\caption{The effect of parameter tuning of IVFADC~\cite{tpami_jegou2011} with 64-bit codes. Left: the runtime per query for SIFT1B dataset, with $w=1$. Right: accuracy and runtime with various $w$ for $\#cell=1024$ and $N=10^8$. }
	\label{fig:teaser_ivfadc}
\end{figure}

However, such inverted indexing systems are built by a process of carefully designed manual parameter tuning, which imply that runtime and accuracy strongly depend on the selection of parameters.
We show the two examples of the effect of such parameter selection
in \Fref{fig:teaser_ivfadc},
using an inverted file with Asymmetric Distance Computation (IVFADC)~\cite{tpami_jegou2011}.
\begin{itemize}
	\item The left figure shows the runtime over the number of database vectors $N$
	with various number of cells ($\#cell$).
	The result with smaller $\#cell$ is faster for $N=10^6$, but
	that with larger $\#cell$ is faster for $N=10^9$.
	Moreover, the relationship is unclear for $10^6<N<10^9$.
	These unpredictable phenomena do not become clear until the searches with several $\#cell$ are examined; however, testing the system is computationally expensive.
	For example, to plot a single dot of \Fref{fig:teaser_ivfadc} for $N=10^9$,
	training and building the index structure took around four days in total. 
	This is particularly critical for recent per-cell training methods~\cite{corr_babenko2014, cvpr_kalantidis2014},
	which require even more computation to build the system.
	\item Another parameter-dependency is given in \Fref{fig:teaser_ivfadc}, right, where
	the runtime and the accuracy in the search range $w$ are presented.
	It has been noted in the existing literature that, with larger $w$, slower but more accurate searches are achieved.
	However, this relation is not simple.
	When compared the result with $w=1$ and $w=8$, the relationship is preserved.
	However, the accuracy with $w=64$ is almost identical to that of $w=8$ even though the search is eight times slower.
	This result implies that users might perform the search with the same accuracy
	but several times slower if they fail to tune the parameter.
\end{itemize}
These results confirm that achieving state-of-the-art performances
depends largely on special tuning for the testbed dataset such as SIFT1B~\cite{icassp_jegou2011}.
In such datasets, recall rates can be easily examined as the ground truth results are given; this is not always true for real-world data.
There is no guarantee of achieving the best performance with such systems.
In real-world applications,
cumbersome trial-and-error-based parameter tuning is often required.

To achieve an ANN system that would be suitable for practical applications,
we propose a \textit{PQTable}; an exact, non-exhaustive, NN search method for PQ codes (see \Fref{fig:teaser_fig}).
We do not employ an inverted index data structure, but find similar PQ codes directly from a database.
This achieves the same accuracy as a linear ADC scan, but requires significantly less time.
(6.9 ms, instead of 8.4 s, for the SIFT1B data using 64-bit codes).
In other words, this paper proposes an efficient ANN search scheme to replace a linear ADC scan when $N$ is sufficiently large.
As discussed in \Sref{sec:anal}, the parameter values required to build the PQTable can be calculated automatically.

The main characteristic of the PQTable is the use of a hash table (see \Fref{fig:teaser_fig}).
An item identifier is associated with the hash table by using its PQ code as a key.
In the querying phase, a query vector is first PQ encoded, and identifiers associated with the key are then retrieved.


A preliminary version of this work appeared in our recent conference paper~\cite{iccv_matsui2015}. This paper contains the following significant differences:
(1) we improved the table-merging step; 
(2) the analysis of collision was provided; 
(3) Optimized Product Quantization~\cite{tpami_ge2014} was incorporated;
and (4) we added massive experimental evaluation using Deep1B dataset~\cite{cvpr_babenko2016}.

The rest of the paper is organized as follows: \Sref{sec:related} introduces related work.
\Sref{sec:pq} briefly reviews the product quantization. 
\Sref{sec:pqtable} presents our proposed PQTable,
and \Sref{sec:anal} shows an analysis for parameter selection.
Experimental results and extensions to Optimized Product Quantization are given in \Sref{sec:exp} and \Sref{sec:extension_opqtable}, respectively.
\Sref{sec:conclusion} presents our conclusions.

\section{Related work}
\label{sec:related}

\subsection{Extensions to PQ}
Since PQ was originally proposed,
several extensions have been studied.
Optimized product quantization (OPQ)~\cite{cvpr_norouzi2013, tpami_ge2014} rotates an input space 
to minimize the encoding error.
Because OPQ always improves the accuracy of encoding with just an additional matrix multiplication,
OPQ has been widely used for several tasks.
Our proposed PQTable can naturally handle OPQ codes. We present the results with OPQ in \Sref{sec:exp_opqtable}.

Additive quantization~\cite{cvpr_babenko2014, eccv_martinez2016} and composite quantization~\cite{icml_zhang2014, cvpr_zhang2015}
generalize the representation of PQ 
from the concatenation of sub-codewords
to the sum of full-dimensional codewords.
These generalized PQs are more accurate than OPQ;
however, they require a more complex query algorithm.

In addition, recent advances of PQ-based methods include novel problem settings such as
supervised~\cite{cvpr_wang2016} and multi-modal~\cite{cvpr_zhang2016}.
Hardware-based acceleration is discussed as well, including GPU~\cite{cvpr_wieschollek2016, corr_johnson2017} and cache-efficiency~\cite{vldb_andre2015}.

\subsection{Hamming-based ANN methods}
As an alternative to PQ-based methods, another major approach to ANN are Hamming-based methods~\cite{corr_wang2016, ieee_wang2015},
in which two vectors are converted to bit strings whose Hamming distance approximates their Euclidean distance.
Comparing bit strings is faster than comparing PQ codes, but is usually less accurate for a given code length~\cite{cvpr_he2013}. 

In Hamming-based approaches, bit strings can be linearly scanned by comparing their Hamming distance,
which is similar to linear ADC scanning in PQ.
In addition, to facilitate a fast, non-exhaustive ANN search, a multi-table algorithm has been proposed~\cite{tpami_norouzi2014}.
Such a multi-table algorithm makes use of hash-tables, where a bit-string itself is used as a key for the tables.
The results of the multi-table algorithm are the same as those of a linear Hamming scan, but the computation is much faster.
For short codes, a more efficient multi-table algorithm was proposed~\cite{tmm_song2016}, and these methods were then extended to the approximated Hamming distance~\cite{cvpr_ong2016}.

Contrarily, a similar querying algorithm and data structure for the PQ-based method
has not been proposed to date.
Our work therefore extends the idea of these multi-table algorithms to the domain of PQ codes,
where a Hamming-based formulation cannot be directly applied. 
\Tref{tbl:hamming_pq_cmp} summarizes the relations between these methods.

The connection between PQ-based and Hamming-based methods is also discussed, including polysemous codes~\cite{eccv_douze2016},
k-means hashing~\cite{cvpr_he2013}, and distance-table analysis.~\cite{mm_wang2014}

\begin{table}
	\begin{center}
		\caption{Relation among Hamming-based and PQ-based ANN methods.}
		\begin{tabular}{@{}lll@{}} \toprule
			& \multicolumn{2}{c}{Search algorithm} \\ \cmidrule(l){2-3}
			& Exhaustive & Non-exhaustive \\ \midrule
			Hamming-based & Linear Hamming scan & Multi-table~\cite{tpami_norouzi2014, tmm_song2016, cvpr_ong2016} \\
			PQ-based & Linear ADC scan & \textbf{PQTable (proposed)} \\ \bottomrule
		\end{tabular}
		\label{tbl:hamming_pq_cmp}
	\end{center}
\end{table}

\section{Background: Product Quantization}
\label{sec:pq}
In this section, we briefly review the encoding algorithm and search process of product quantization~\cite{tpami_jegou2011}.

\subsection{Product quantizer}
Let us denote any $D$-dimensional vector $\bvec{x} \in \mathbb{R}^D$ as a concatenation of $M$ subvectors: $\bvec{x} = [(\bvec{x}^1)^\top, \dots, (\bvec{x}^M)^\top]^\top$,
where each $\bvec{x}^m \in \mathbb{R}^{D/M}$. We assume $D$ can be divided by $M$ for simplicity.
A product quantizer, $\mathbb{R}^D \to \{1, \dots, K\}^M$ is defined as follows\footnote{
	In this paper, we use a bold font to represent vectors.
	A square bracket with a non-bold font indicates
	an element of a vector.
	For example, given $\bvec{a}\in \mathbb{R}^D$, $d$th element of $\bvec{a}$ is $a[d]$, i.e.,
	$\bvec{a}=[a[1], \dots, a[D]]^\top$.
}:
\begin{equation}
\bvec{x} \mapsto \bvec{\bar{x}} = [\bar{x}[1], \dots, \bar{x}[M]]^\top.
\end{equation}
Each $\bar{x}[m]$ is a result of a subquantizer: $\mathbb{R}^{D/M} \to \{1, \dots, K \}$ defined as follow:
\begin{equation}
\mathbf{x}^m \mapsto \bar{x}[m] = \argmin_{k \in \{1, \dots, K\}} \Vert \mathbf{x}^m - \mathbf{c}_k^m\Vert_2^2.
\end{equation}
Note that $K$ $D/M$-dim codewords $C^m=\{\mathbf{c}_k^m\}_{k=1}^K, \mathbf{c}_k^m \in \mathbb{R}^{D/M}$ are trained for each $m$ in advance by k-means~\cite{tit_lloyd1982}.
In summary, the product quantizer divides an input vector into $M$ subvectors, quantizes each subvector to an integer ($1, \dots, K$), and concatenates resultant $M$ integers.
In this paper, this product-quantization is also called ``encoding'', and a bar-notation ($\bvec{\bar{x}}$) is used to represent a PQ code of $\bvec{x}$.

A PQ code is represented by $B=M\log_2 K$ bits.
Typically, $K$ is set as a power of 2, making $\log_2 K$ an integer.
In this paper, we set $K$ as 256 so that $B=8M$. 

\subsection{Asymmetric distance computation}
Distances between a raw vector and a PQ code can be approximated efficiently.
Suppose that there are $N$ data points, $\mathcal{X}=\{\bvec{x}_n\}_{n=1}^N$, and they are PQ-encoded as a set of PQ codes
$\mathcal{\bar{X}} = \{\bvec{\bar{x}}_n\}_{n=1}^N$.
Given a new query vector $\bvec{q} \in \mathbb{R}^D$, the squared Euclidean distance
from $\bvec{q}$ to $\bvec{x}\in\mathcal{X}$ is approximated using the PQ code $\bvec{\bar{x}}$.
This is called an asymmetric distance (AD)~\cite{tpami_jegou2011}:
\begin{equation}
d(\bvec{q}, \bvec{x})^2 \sim d_{AD}(\bvec{q}, \bvec{x})^2 = \sum_{m=1}^M d\left(\mathbf{q}^m, \mathbf{c}^m_{\bar{x}[m]}\right)^2 
\label{eq:adc}
\end{equation}
This is computed as follows:
First, $\bvec{q}^m$ is compared to each $\bvec{c}_k^m \in \mathcal{C}^m$, thereby generating a distance matrix online,
where its $(m, k)$ entry denotes the squared Euclidean distance between $\bvec{x}^m$ and $\bvec{c}^m_k$.
For the PQ code $\mathbf{\bar{x}}$, the decoded vector for each $m$ is fetched as $\mathbf{c}^m_{\bar{x}[m]}$.
The $d_{AD}$ approximates the distance between the query and the original vector $\bvec{x}$
using the distance between the query and this decoded vector.
Furthermore, the computation can be achieved by simply looking up the distance matrix 
($M$ times checking and summing).
The computational cost for all $N$ PQ-codes is $O(DK+MN)$, which is fast for small $N$ but still linear in $N$.

\section{PQTable}
\label{sec:pqtable}

\subsection{Overview}

\begin{figure*}
	\begin{center}
		\subfloat[Single table ($M=4$, $B=32$, and $L=1$). The number of slots is $(256)^M=2^{32} \sim 4.3\times10^9$.]{\includegraphics[width=0.35\linewidth]{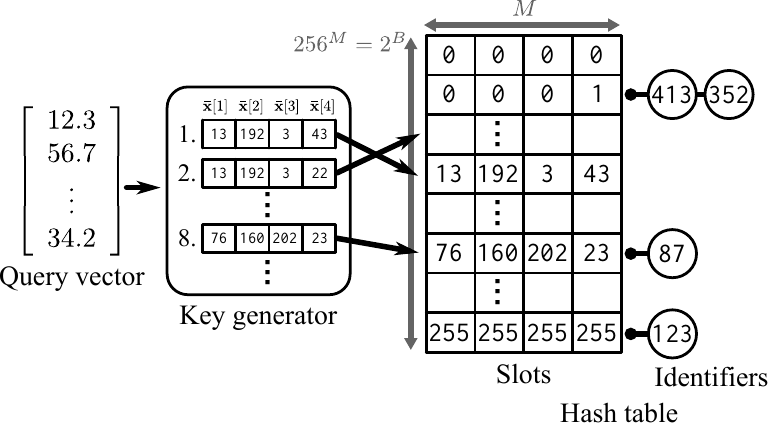}
			\label{fig:overview1}}
		~~~~
		\subfloat[Multitables ($M=4$, $B=32$, $T=2$, and $L=2$). The number of slots for each table is $256^{M/T}=2^{16}\sim6.6\times10^4$.]{\includegraphics[width=0.6\linewidth]{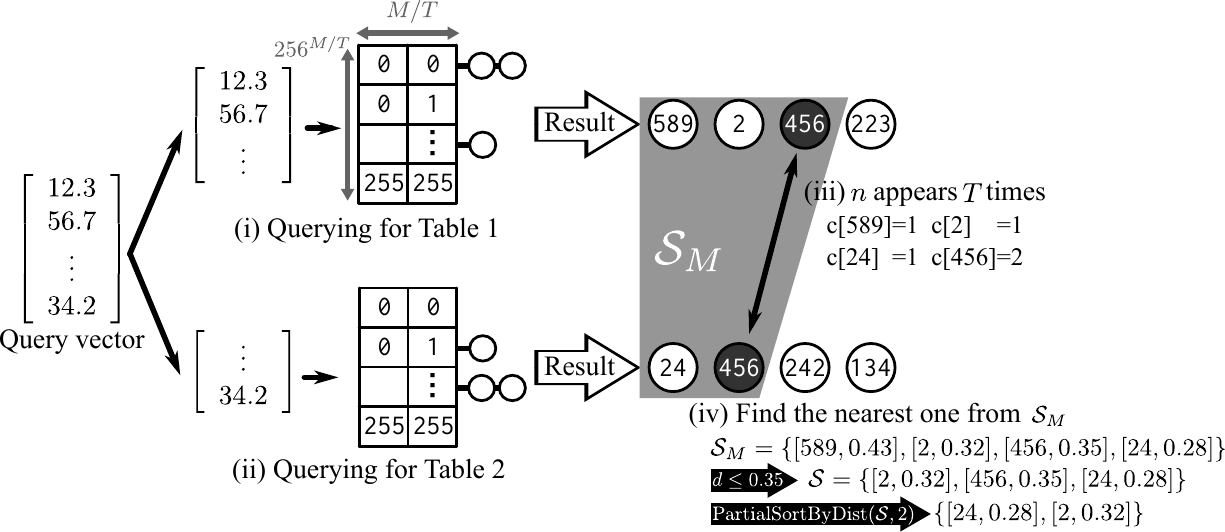}
			\label{fig:overview2}}
	\end{center}
	\caption{Overview of the proposed method.}
	\label{fig:overview}
\end{figure*}

In this section, we introduce the proposed PQTable.
As shown in \Fref{fig:teaser_fig}, the basic idea of the PQTable is to use a hash table.
Each slot of the hash table is a concatenation of $M$ integers. 
For each slot, a list of identifiers is associated.
In the offline phase, given a $n$th database item (a PQ code $\bvec{\bar{x}}_n$ and an identifier $n$),
the PQ code $\bvec{\bar{x}}_n$ itself is used as a key.
The identifier $n$ is inserted in the slot. 
In the retrieval phase, a query vector is PQ-encoded to create a key.
Identifiers associated with the key are retrieved.
Accessing the slot (i.e., hashing) is an $O(1)$ operation.
This process seems very straightforward, but there are two problems to be solved.

\subsubsection{The empty-entries problem}
Suppose a new query has been PQ encoded and hashed.
If the identifiers associated with the slot are not present in the table,
the hashing fails. To continue the retrieval, we would need to find new candidates by some other means.
To handle this empty-entries problem, we present a \textbf{key generator},
which is mathematically equivalent to a multi-sequence algorithm~\cite{cvpr_babenko2012}.
This generator creates next nearest candidates one by one, as shown in \Fref{fig:overview1}.
For a given query vector, the generator produces the first nearest code $\mathbf{\bar{x}}=[13, 192, 3, 43]^\top$, which is then hashed;
however, the table does not contain identifiers associated with that code.
The key generator then creates and hashes a next nearest code $\mathbf{\bar{x}}=[13, 192, 3, 22]^\top$.
In this case, we can find the nearest identifier (``87'') at the eighth-time hashing.

\subsubsection{The long-code problem}
Even if we can find candidates and continue querying,
the retrieval is not efficient if the length of the codes is long compared to the number of slots,
e.g., $B=64$ codes for $N=10^9$ vectors.
Let us recall the example of \Fref{fig:overview1}.
The number of slots (the size of the hash table) is $K^M=256^M=2^B$,
whereas the sum of the number of the identifiers is $N$.
If $N$ is much smaller than $2^B$, the hash table becomes sparse (almost all slots will be empty), and this results in an inefficient querying as we cannot find any identifiers, even if a large number of candidates are created.
To solve the long-code problem, we propose \textbf{table division and merging}, as shown in \Fref{fig:overview2}.
The hash table is divided into $T$ small hash tables. Querying is performed for each table, and the results are merged.

We first show the data structure and the querying algorithm for a single table, which uses the key generator to solve the empty entries problem (\Sref{sec:single_hash_table}).
Next, we extend the system to multiple hash tables using table division and merging to overcome the long-code problem (\Sref{sec:multi_hash_table}).

\subsection{Single Hash Table}
\label{sec:single_hash_table}

First, we show a single-table version of the PQTable (\Fref{fig:overview1}).
The single PQTable is effective when
the difference between $N$ and $2^B$ is not significant.
A pseudo-code\footnote{
	If we speak in a \texttt{c++} manner, \Aref{alg:single_table} shows a definition of the PQTable \texttt{class}.
	The elements in \textbf{Member} are member variables. \textbf{Function}s are regarded as member functions.
} is presented in \Aref{alg:single_table}.
A PQTable is instantiated with a hash-table $tbl$ and a key generator $keygen$ (L2-L3 in \Aref{alg:single_table}).
We give the pseudocode of our implementation of the key generator in \Appref{sec:multisequence} as a reference.

\subsubsection{Offline}
The offline step is described in \texttt{Insert} function (L4-L6).
In the offline step, database vectors $\{\bvec{x}_n\}_{n=1}^N$ are PQ-encoded first.
The resultant PQ codes $\{\bvec{\bar{x}}_n\}_{n=1}^N$ are inserted into a hash table (L6).
The function \texttt{Push}$(\bvec{\bar{x}}, n)$ of $tbl$ means inserting an identifier $n$ to $tbl$ 
using $\bvec{\bar{x}}$ as a key.
If identifiers already exist in the slot, the new $n$ is simply added to the end
(e.g., ``413'' and ``352'' are associated with the same slot $[0, 0, 0, 1]^\top$ in \Fref{fig:overview1})

\begin{algorithm}[t]
	\SetKwProg{Fn}{Function}{}{End} 
	\SetKwProg{Mem}{Member}{}{End} 
	\Mem{}{
		$tbl \gets \emptyset$ ~~~~~~~~~~~~~~~~~~~~~~ \codecomment{// Hash-table} \\
		$keygen \gets KeyGenerator$ \codecomment{// Key generator is instantiated (\Aref{alg:msa})}
	}
	\Fn{$\mathrm{Insert}$}{
		\KwIn{$\{ \bvec{\bar{x}}_n \}_{n=1}^N,~\bvec{\bar{x}}_n\in\{1, \dots, K\}^M$\codecomment{// PQ-codes}}
		\For{$n \gets 1~\mathrm{to}~N$}{
			$tbl$.Push($\bvec{\bar{x}}_n$, $n$)
		}			
	}
	\Fn{$\mathrm{Query}$}{
		\KwIn{$\bvec{q}\in\mathbb{R}^D$,~~~~~~~~~~~~~\codecomment{// Query vector} \\
			~~~~~~~~~$L\in\{1, \dots, N\}$.~~~\codecomment{// \#returned items}}
		\KwOut{$\mathcal{S}=\{ \mathbf{s}_l \}_{l=1}^L,~\mathbf{s}_l = [n_l, d_l] \in \{1, \dots, N \}\times\mathbb{R}$\\~~~~~~~~~~~\codecomment{// Top $L$ smallest scores}
		}
		$\mathcal{S} \gets \emptyset$ \\
		$keygen.$Init$(\mathbf{q})$ \\
		\While{$|\mathcal{S}|<L$}{
			$\bvec{\bar{x}},~d \gets keygen.$NextKey() \\
			$\{n_1, n_2, n_3, \dots \} \gets tbl.$Hash$(\bvec{\bar{x}})$ \\
			\ForEach{$n \in \{ n_1, n_2, n_3, \dots\}$}{
				$\mathcal{S} \gets \mathcal{S} \cup [n, d]$~~\codecomment{// Push back} 
			}
		}
		\KwRet{$\mathcal{S}$}
	}
	
	\caption{Single PQTable}
	\label{alg:single_table}
\end{algorithm}

\subsubsection{Online}
The online step is presented in \texttt
{Query} function (L7-L15).
In the online step, the function takes a query vector $\bvec{q}$ and the length of the returned item $L$ as inputs.
The function then retrieves $L$ nearest items ($L$ pairs of an identifier and a distance).
We denote these nearest items as $\mathcal{S}=\{\bvec{s}_l\}_{l=1}^L$,
where each $\bvec{s}_l$ is a pair of two scholar values.

First, the key generator is initialized using $\bvec{q}$ (L9), and the search continues until the $L$ items are retrieved (L10).
For each loop, the next nearest PQ code $\bvec{\bar{x}}$ (and AD $d$) is created (L11).
This iterative creation is visualized in the ``Key generator'' box in \Fref{fig:overview1}.
When the \texttt{NextKey} function of $keygen$ is called,
the next nearest PQ code $\bvec{\bar{x}}$ (in terms of AD from the query) is returned.
Using $\bvec{\bar{x}}$ as a key, the associated identifiers are found from $tbl$ (L12).
The \texttt{Hash} function returns all identifiers ($n_1, n_2, \dots, $) associated with the slot.
For example, $n_1=413$ and $n_2=352$ are returned if the key $[0, 0, 0, 1]^\top$ is hashed in \Fref{fig:overview1}. 
For each $n \in \{n_1, n_2, \dots \}$, the code and the distance is pushed into $\mathcal{S}$.
Owing to \texttt{NextKey}, we can continue querying even if identifiers with the focusing slot are empty, thereby solving the empty-entries problem.

Note again that $keygen$ is mathematically equivalent to the higher-order
multi-sequence algorithm~\cite{cvpr_babenko2012}, which was originally used to divide the space into Cartesian products for coarse quantization.
We found that it can be used to enumerate PQ code combinations
in the ascending order of AD.

If sufficient numbers of items are collected, 
items $\mathcal{S}$ are returned (L15).
Note that, if $L$ is small enough, 
the table immediately returns $\mathcal{S}$; i.e., the first $n_1$ is returned without fetching $\{n_2, n_3, \dots \}$ if $L$ is one. This accelerates the performance.

\subsection{Multiple Hash Table}
\label{sec:multi_hash_table}

\begin{algorithm}[t]
	\SetKwProg{Fn}{Function}{}{End} 
	\SetKwProg{Mem}{Member}{}{End} 
	\SetKwFor{InfRepeat}{Repeat}{}{} 
	\Mem{}{
		$\{{tbl}_1, \dots, {tbl}_T\}$, where each ${tbl}_t \gets \emptyset$ ~~~~~~~~~~~~~~~~~~~\codecomment{// $T$ small Hash-tables} \\
		$\{{keygen}_1, \dots, {keygen}_T \},$ where each ${keygen}_t \gets KeyGenerator$ \codecomment{{\footnotesize // $T$ KeyGenerators}} \\
		$\mathcal{\bar{X}} \gets \emptyset$ \codecomment{// PQ-codes}
	}
	\Fn{$\mathrm{Insert}$}{
		\KwIn{$\{ \bvec{\bar{x}}_n \}_{n=1}^N,~~~\bvec{\bar{x}}_n\in\{1, \dots, K\}^M$}
		\For{$n \gets 1~\mathrm{to}~N$}{
			\For{$t \gets 1~\mathrm{to}~T$}{
				${tbl}_t$.Push($\left [\bar{x}_n\left[1+\frac{M}{T}(t-1)\right], \dots, \bar{x}_n\left[\frac{M}{T}t\right] \right ]^\top$,~$n$)
			}
		}
		$\mathcal{\bar{X}} \gets \{ \bvec{\bar{x}}_n \}_{n=1}^N$
	}
	\Fn{$\mathrm{Query}$}{
		\KwIn{$\bvec{q}\in\mathbb{R}^D$,~~~~~~~~~~~~~\codecomment{// Query vector} \\
			~~~~~~~~~$L\in\{1, \dots, N\}$.~~~\codecomment{// \#returned items}}
		\KwOut{$\mathcal{S}=\{ \mathbf{s}_l \}_{l=1}^L,~\mathbf{s}_l = [n_l, d_l] \in \{1, \dots, N \}\times\mathbb{R}$\\~~~~~~~~~~\codecomment{// Top $L$ smallest scores}}					
		$\mathcal{S} \gets \emptyset$ \\
		$\mathcal{S}_M \gets \emptyset$~\codecomment{// Marked scores (tmp. buffer)}\\
		$\bvec{c} \gets \emptyset$~~~~\codecomment{// Counter. $\mathbf{c}\in\{0, \dots, T\}^N$}\\ 
		\For{$t \gets 1~\mathrm{to}~T$}{
			${keygen}_t.$Init$(\left[q\left[1+\frac{D}{T}(t-1)\right], \dots, q\left[\frac{D}{T}t\right] \right]^\top)$ \\
		}
		\InfRepeat{}{
			\For{$t \gets 1~\mathrm{to}~T$}{
				$\bvec{\bar{x}} \gets {keygen}_t.$NextKey()\codecomment{{\footnotesize~//~$\bvec{\bar{x}}\in\{1, \dots, K\}^{\frac{M}{T}}$}}\\
				$\{n_1, n_2, n_3, \dots \} \gets {tbl}_t.$Hash$(\bvec{\bar{x}})$ \\
				\ForEach{$n \in \{ n_1, n_2, n_3, \dots\}$}{
					$c[n] \gets c[n] + 1$ \\
					\uIf{$c[n] = 1$}{
						$\mathcal{S}_M \gets \mathcal{S}_M \cup [n, d_{AD}(\bvec{q}, \bvec{x}_n)]$ \\
					}
					\ElseIf{$c[n] = T$}{
						$d_{min} \gets d_{AD}(\bvec{q}, \bvec{x}_n)$ \\
						$\mathcal{S} \gets \{[n, d] \in \mathcal{S}_M | d \le d_{min}  \}$ \\
						\If{$L \le |\mathcal{S}|$}{
							\KwRet{$\mathrm{PartialSortByDist}(\mathcal{S}, L)$} \codecomment{// {\footnotesize $L$ smallest sorted scores}}
						}
					}
					
				}				
			}
		}
		
	}
	\caption{Multiple PQTable}
	\label{alg:multi_table}
\end{algorithm}

The single-table version of the PQTable may not work when the code-length is long, e.g., $64 \le B$.
This is the long code problem described above,
where the number of possible slots ($2^B$) is too large for efficient processing.
For example, in our experiment, the maximum number of database vectors ($N$) is one billion. Therefore, most slots will be empty if $64 \le B$ (i.e., $2^{64} \sim 1.8 \times 10^{19} \gg 10^9$).

To solve this problem, we propose a table division and merging method.
The table is divided as shown in \Fref{fig:overview2}.
If an $B$-bit code table is divided into $T$ $B/T$-bit code tables,
the number of the slots decreases, from $2^B$ for one table to $2^{B/T}$ for $T$ tables.
By properly merging the results from each of the small tables, we can obtain the correct result.

A pseudocode is presented in \Aref{alg:multi_table}. The multi-PQTable is instantiated with
$T$ hash tables and $T$ key generators.

\subsubsection{Offline}
The offline step is described in \texttt{Insert} function (L5-L9).
Each input PQ code $\bvec{\bar{x}}$ is divided into $T$ parts.
$t$th part is used as a key for $t$th $tbl$ to associate an identifier (L8).
For example, if $\bvec{\bar{x}}_{93}=[13, 35, 7, 9]$ and $T=2$, the first part $[13, 35]$ is 
used as a key for the first table ${tbl}_1$, and $n=93$ is inserted.
The second part $[7, 9]$ is used for the second table, then $n=93$ is inserted.
Unlike the single PQTable, the PQ codes themselves are also stored (L9).

\subsubsection{Online}
The online querying step is presented in \texttt{Query} function (L10-L28).
The inputs and outputs are as the same as those for the single PQTable.
In addition to the final scores $\mathcal{S}$, we prepare a temporal buffer $\mathcal{S}_M$ which is also a set of scores called ``marked scores'' (L12).
As a supplemental structure, we prepare a counter $\bvec{c} \in \{0, \dots, T\}^N$ (L13).
The counter counts the frequency of the number $n$.
For example, $c[13]=5$ means that $n=13$ appears five times\footnote{
	We cannot use an array to represent $\mathbf{c}$ for a large $N$, as $\bvec{c}$ could require more memory space than the
	hash table itself.
	We do not need to prepare a full memory space for all $N$
	because only a fraction of $n$s are accessed.
	In our implementation, a standard associative array (\texttt{std::unordered\_map}) is leveraged to represent $\mathbf{c}$.
}.

An input query $\bvec{q}$ is divided into $T$ small vectors,
and $t$th key generator is initialized by the $t$th small vector (L15).
The search is performed in the Repeat loop (L16).
For each $t$, a small PQ-code ($\{1, \dots, K\}^{M/T}$) is created, hashed, and
the associate identifiers are obtained in the same manner as the single PQ-table (L17-L20).
This step means finding similar codes by just seeing the $t$th part.
\Fref{fig:overview2}(i, ii) visualizes these operations,
where the associated identifiers are retrieved for each hash table.
In this case, the 589th PQ code is the nearest to the query if we see only a first half of the vector, but this is not necessarily the case regarding the last half. 

Let us describe the proposed result-merging step.
Given $n \in \{n_1, n_2, \dots\}$, the next step is counting the frequency of $n$; this can be done by simply updating $c[n] \gets c[n] + 1$ (L21).
These identifiers are possible answers of the problem because
at least $t$th part of the PQ code is similar to the query.
When $n$ appears for the first time ($c[n]=1$), 
we compute the actual asymmetric distance between the query and $n$th item ($d_{AD}(\bvec{q}, \bvec{x}_n)$).
This can be achieved by picking up the PQ-code $\bvec{\bar{x}}_n$ from $\mathcal{\bar{X}}$ (L23).
At the same time, we store a pair of $n$ and the computed $d_{AD}$ in $\mathcal{S}_M$.
We call this step ``marking'', as visualized by a gray color in \Fref{fig:overview2}.
The key generation, hashing, and marking are repeated
until we find $n$ such that $n$ appears $T$ times, i.e., $c[n]=T$ (L24).
Let us denote the $d_{AD}$ of this $n$ as $d_{min}$ (L25).
It is guaranteed that \textbf{any items whose $d_{AD}$ is less than $d_{min}$ are already marked}
(see \Appref{sec:proof} for the proof).
Therefore, the final set of scores $\mathcal{S}$ can be
constructed by picking up items whose $d_{AD}$ is less than $d_{min}$ from the marked items (L26).
If the number of the scores $|\mathcal{S}|$ is more than the required number $L$,
the scores are sorted partially and the top $L$ scores are returned (L27-L28).
\Fref{fig:overview2}(iii, iv) visualizes these processes.
Here, we find that $n=456$ appears $T=2$ times (iii), so we evaluate the items in $\mathcal{S}_M$.
The $d_{min}$ is 0.35, so the items that have $d\le0.35$ are selected to construct $\mathcal{S}$.
In this case, a sufficient number ($L=2$) of items are in $\mathcal{S}$.
Thus, the items in $\mathcal{S}_M$ are partially sorted, and the top-$L$ results are returned.
If there are not enough items in $\mathcal{S}$, entire loop continues until a sufficient number of items is found.

The intuition of the proposed merging step is simple;
marked items have a high possibility of being the nearest items,
but much closer PQ code might exist which may be unmarked.
If we can find a ``bound'' of the distance
where any items that are closer than the bound must be marked,
we simply need to evaluate the marked items.
The item which appears $T$ times acts as this bound.

\subsubsection{Implementation details}
Although we use two sets $\mathcal{S}$ and $\mathcal{S}_M$ in \Aref{alg:multi_table} for the ease of explanation,
they can be implemented by maintaining an array.
In the case $L=1$, we simplify L25-L28 as we do not need to sort results.

\section{Analysis for Parameter Selection}
\label{sec:anal}
In this section, we discuss how to determine the value of the parameter required to construct the PQTable.
Suppose that we have $N$ $B$-bit PQ-codes ($\{\bvec{\bar{x}}_n\}_{n=1}^N, \bvec{\bar{x}}_n\in\{1, \dots, K\}^M, B=M\log K=8M$).
To construct the PQTable, we must select one parameter value $T$ (the number of dividing tables).
If $B=32$, for example, we need to select the data structure as being either a single 32-bit table,
two 16-bit tables, or four 8-bit tables, corresponding to $T=1, 2,$ and $4$, respectively.
To analyze the performance of the proposed PQTable,
we first consider the case in which PQ codes are uniformly
distributed.
Next, we show that the behavior of hash tables is
strongly influenced by the distribution of the database vectors.
Taking this into account, we present an indicative value, $T^*=B / \log_2 N$, as proposed by previous work on multi-table hashing~\cite{focs_greene1994, tpami_norouzi2014}.
We found that this indicative value estimates the optimal $T$ well.

\subsection{Observation}
Considering a hash table, there is a strong relationship between $N$,
the number of slots $2^B$, and the computational cost.
If $N$ is too small, almost all slots will not be associated with identifiers, and generating candidates will take time.
If $N$ is the appropriate size and the slots are well filled, search speed is high.
If $N$ is too large compared with the size of the slots,
all slots are filled and the number of identifiers associated with each slot is large, which can cause slow fetching.

\begin{figure}
	\begin{center}
		\subfloat[32-bit PQ codes from the SIFT1B data.]{\includegraphics[width=1.0\linewidth]{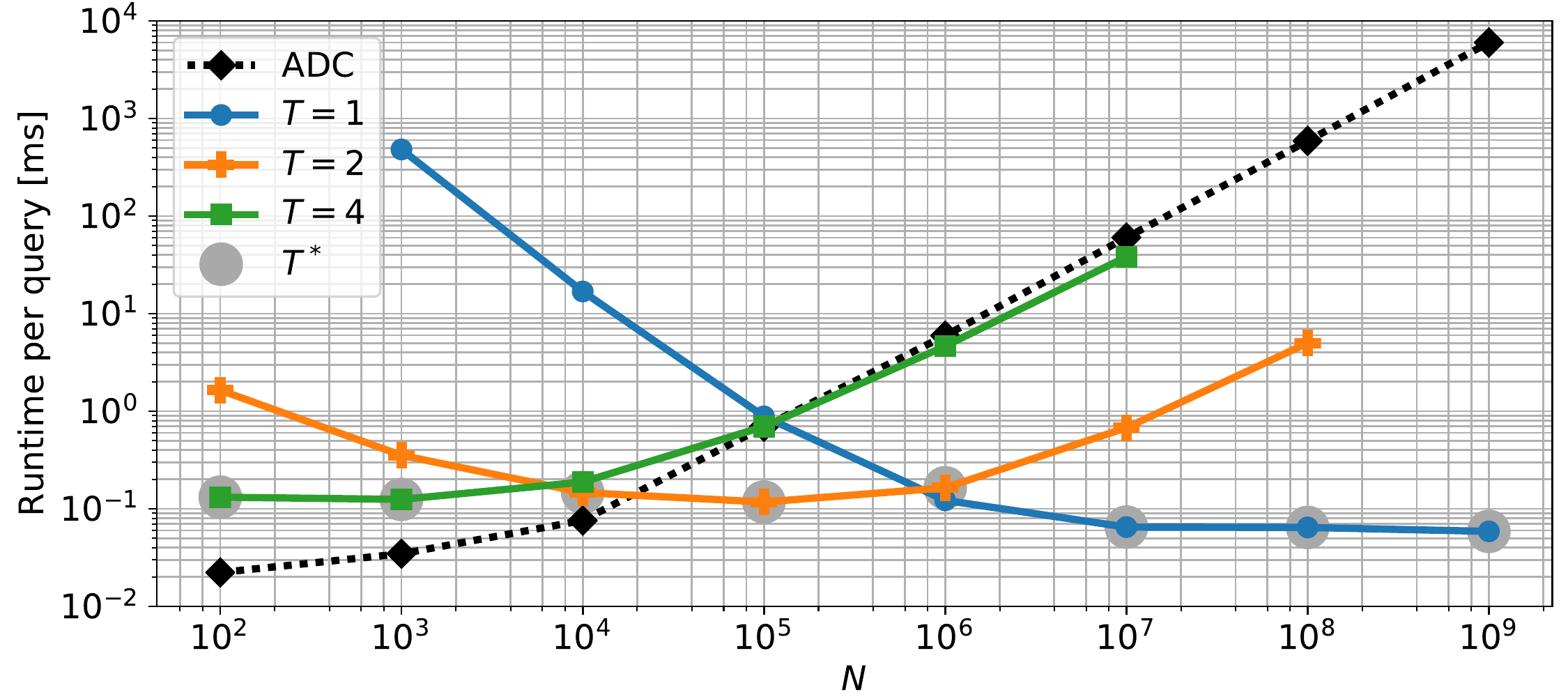}
			\label{fig:anal1}}
		
		\subfloat[64-bit PQ codes from the SIFT1B data.]{\includegraphics[width=1.0\linewidth]{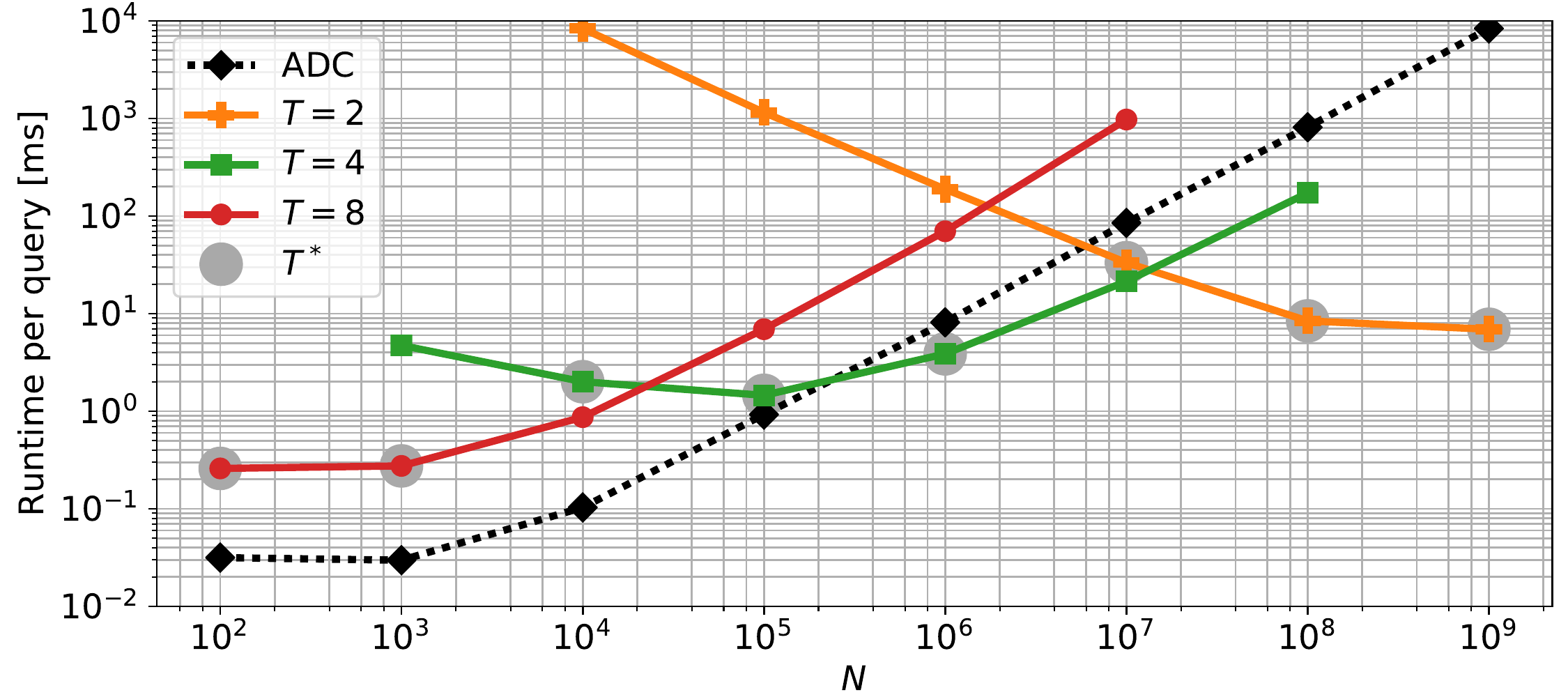}
			\label{fig:anal2}}
	\end{center}
	\caption{Runtime per query of each table.}
	\label{fig:anal}
\end{figure}

\Fref{fig:anal1} shows the relationship between $N$ and the computational time for 32-bit codes with $T=1,2,$ and $4$.
\Fref{fig:anal2} shows that for 64-bit codes and $T=2,4,$ and $8$.
We can find that each table has a ``hot spot.''
In \Fref{fig:anal1},
for $10^2\le N \le 10^3$, $10^4 \le N \le 10^5$, and $10^6 \le N \le 10^9$,
$T=4, 2,$ and $1$ are the fastest, respectively.
Given $N$ and $B$, our objective here is to decide the optimal $T$ without constructing tables.

\subsection{Comparison to uniform distribution}

\begin{figure*}
	\begin{center}
		\subfloat[]{\includegraphics[width=0.33\linewidth]{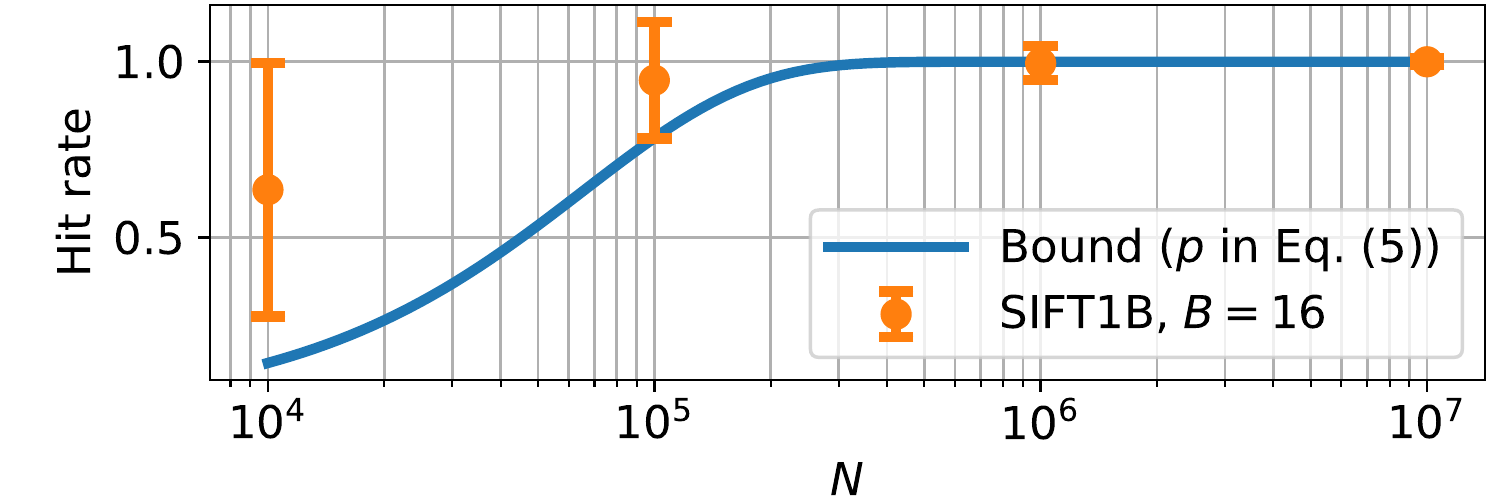}
			\label{fig:anal_unif_p_b16}}		
		\subfloat[]{\includegraphics[width=0.33\linewidth]{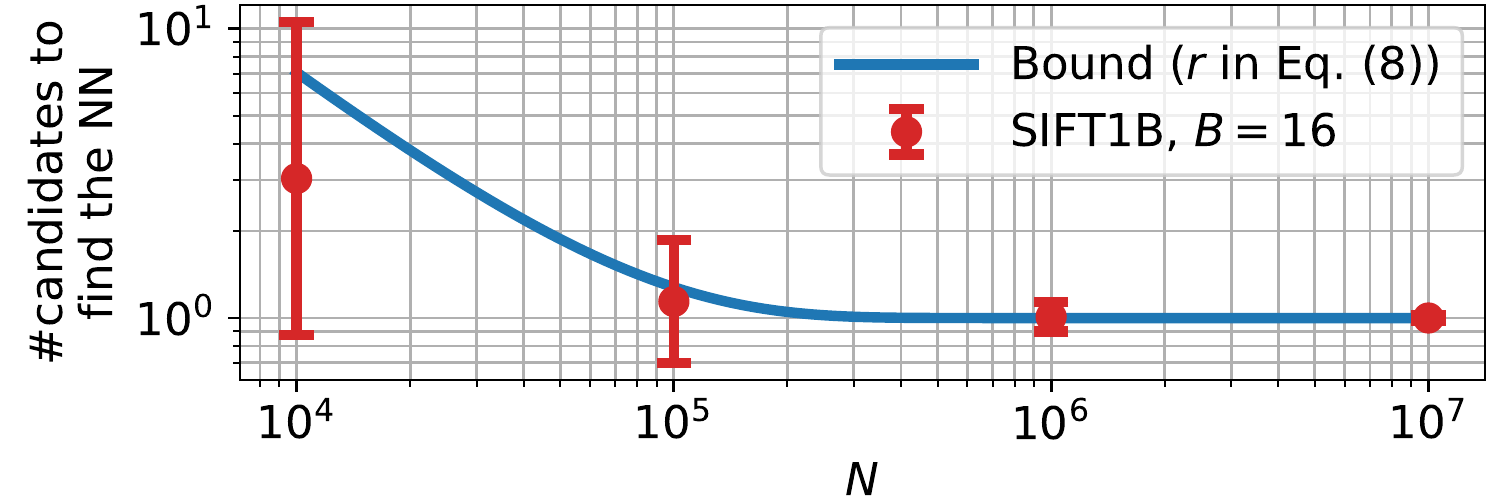}
			\label{fig:anal_unif_r_b16}}
		\subfloat[]{\includegraphics[width=0.33\linewidth]{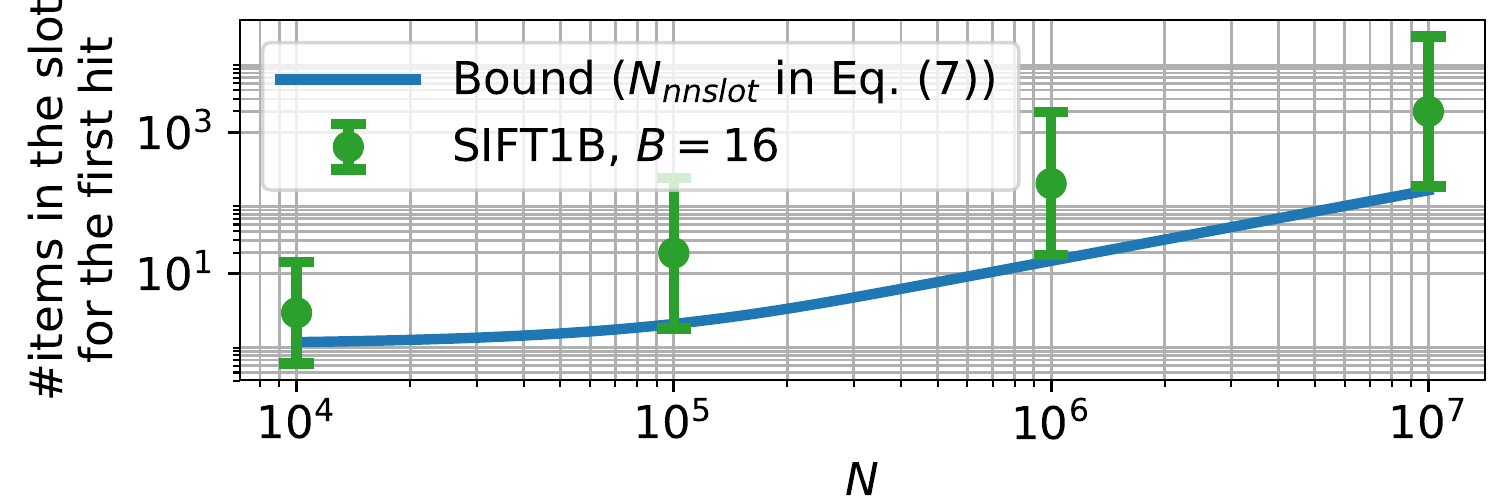}
			\label{fig:anal_unif_nhash_b16}}
		
		\subfloat[]{\includegraphics[width=0.33\linewidth]{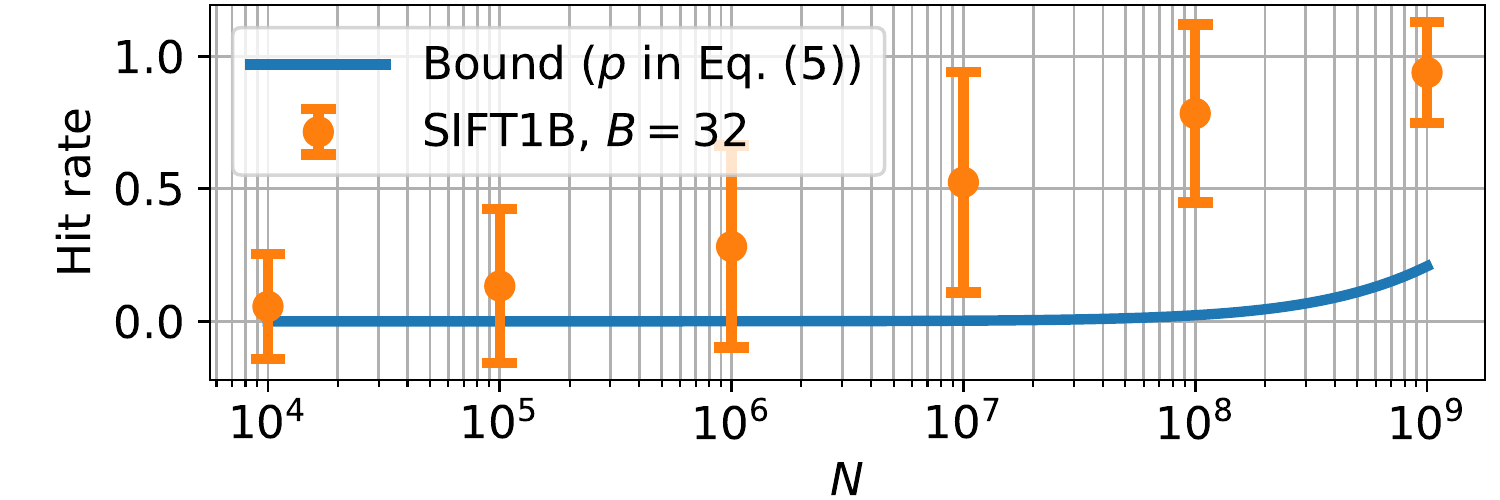}
			\label{fig:anal_unif_p_b32}}		
		\subfloat[]{\includegraphics[width=0.33\linewidth]{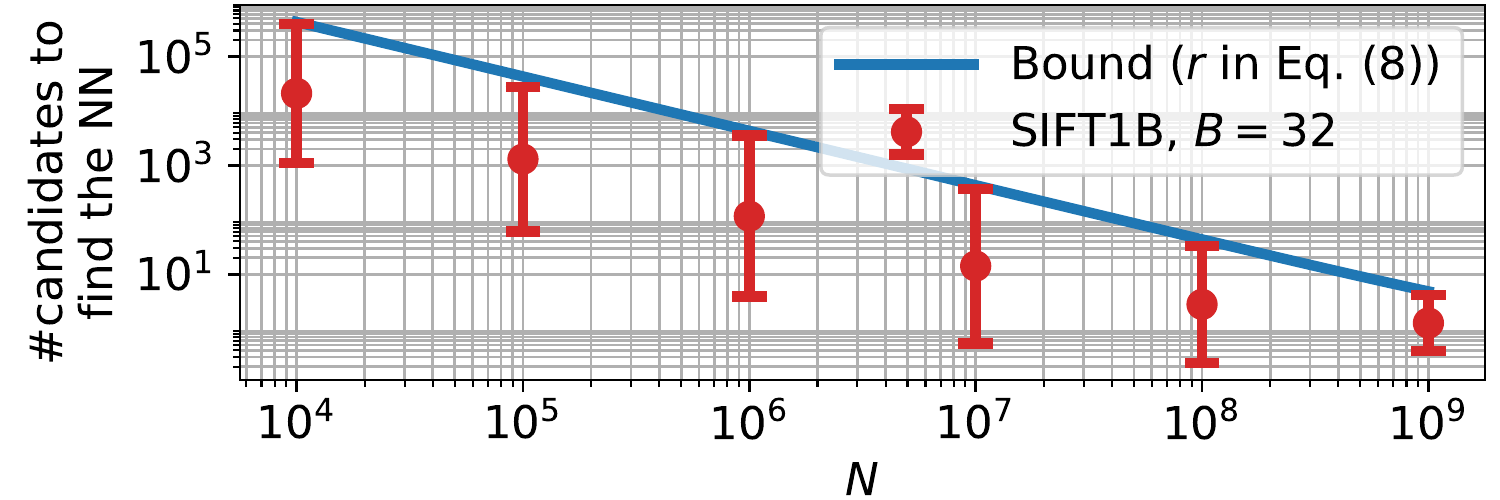}
			\label{fig:anal_unif_r_b32}}
		\subfloat[]{\includegraphics[width=0.33\linewidth]{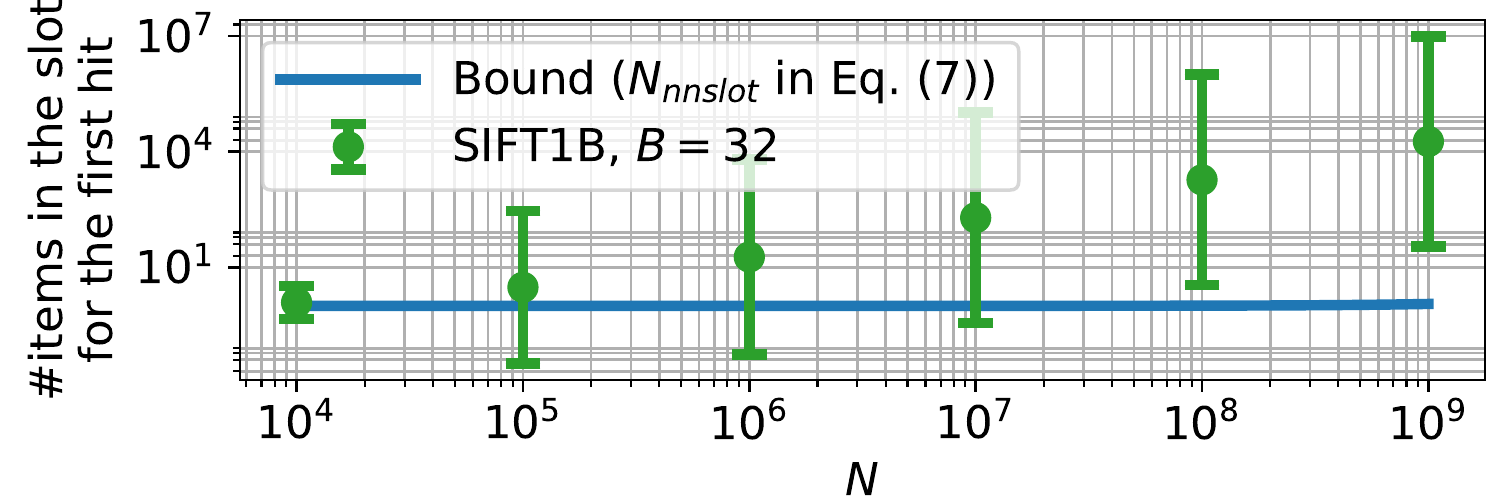}
			\label{fig:anal_unif_nhash_b32}}
	\end{center}
	\caption{Performance analysis using a SIFT1B dataset.
		All results are from a single PQTable ($T=1$).
		Error bar shows the standard deviation over $10^4$ queries.}
	\label{fig:anal_unif}
\end{figure*}

Let us first analyze the case when all items are uniformly distributed in a hash table.
Next, we show that the observed behavior of the items is extremely different. 

We will consider a single hash table for simplicity.
Suppose that each item has an equal probability of hashing to each slot.
First, we focus a fill-rate as follows:
\begin{equation}
p=\frac{\mathrm{\#filled\_slots}}{\mathrm{\#slots}} = \frac{\mathrm{\#slots} - \mathrm{\#empty\_slots}}{\mathrm{\#slots}},
\end{equation}
where $0 < p \le 1$.
Because we consider a $B$-bit hash table, $\mathrm{\#slots} = 2^B$.
The number of the expected value of empty slots is computed as follows:
Because items are equally distributed,
the probability that an entry is empty after we insert an item into the table is $1-\frac{1}{2^B}$. 
Because each hashing can be considered as an independent trial,
the probability of nothing being hashed to a slot in $N$ trials is $(1-\frac{1}{2^B})^N$.
This indicates that the expected value for an entry being empty
(0 for being empty, and 1 for being non-empty) after $N$ insertions
is $(1-\frac{1}{2^B})^N$.
Because of the linearity of the expected value, the expected number of empty slots is
$2^B(1-\frac{1}{2^B})^N$ (Theorem 5.14 in \cite{book_stein2010}).
From this, the fill-rate is denoted as:
\begin{equation}
p = \frac{2^B - 2^B(1-\frac{1}{2^B})^N}{2^B}=1- \left (1-\frac{1}{2^B} \right )^N.
\end{equation}
Note that, for a given query, the probability of the slot being filled is also the same as $p$, as we assume that all queries are also uniformly distributed. We call this probability the hit rate.

Next, we compute $r$, which is the expected number of hashings to find the nearest item.
This value corresponds to the number of iterations of loop L10 in \Aref{alg:single_table}
Because the probability of finding the nearest item for the first time in $r'$th step is
$(1-p)^{r'-1}p$, the expected value of $r'$ is:
\begin{equation}
r=\sum_{r'=1}^\infty (1-p)^{r'-1}pr' = \frac{1}{p} = \frac{1}{1- \left (1-\frac{1}{2^B} \right )^N}.
\end{equation}

Finally, we compute $N_{nnslot}$, which indicates the number of items assigned to the slot when hashing is successful.
This value corresponds to the number of returned items $|\{n_1, n_2, \dots\}|$ in L12 in \Aref{alg:single_table}.
Under uniform distribution, we can count $N_{nnslot}$ by simply dividing the total number $N$ by $\mathrm{\#filled\_slots}$:
\begin{equation}
N_{nnslot} = \frac{N}{\mathrm{\#filled\_slots}} = \frac{N}{2^B(1-(1-\frac{1}{2^B})^N)}.
\end{equation}

\Fref{fig:anal_unif_p_b16} and \Fref{fig:anal_unif_p_b32} show hit rates
for $B=16$ and $B=32$, respectively.
In addition,  we show the observed values of hit-rate over the SIFT1B dataset.
Because the actual data follow some distribution in both
the query the database sides, the hit-rate is higher than $p$;
i.e., $p$ it acts as a bound. 
Similarly, the number of candidates to find the nearest neighbor
is shown in \Fref{fig:anal_unif_r_b16} and \Fref{fig:anal_unif_r_b32},
and the number of items assigned to the slot for the first hit is presented
in \Fref{fig:anal_unif_nhash_b16} and \Fref{fig:anal_unif_nhash_b32}.
Note that $r$ and $N_{nnslot}$ also act as bounds.

Remarkably, the observed SIFT1B data behaves in a completely different way compared to that of the equally-distributed bound, especially for a large $N$.
For example, the observed hit rate for $N=10^9$ is 0.94 in \Fref{fig:anal_unif_p_b32}, whereas the hit rate for $p$ is just 0.21.
Therefore, the first hashing almost always succeeds. 
This is supported in \Fref{fig:anal_unif_r_b32}, where 
the required number for finding the searched item is just 1.3 for $N=10^9$.
At the same time, the number of items assigned to the slot is surprisingly larger than the bound ($1.6\times10^4$ times, as shown in \Fref{fig:anal_unif_nhash_b32}).
This heavily biased behavior is attributed to the distribution of the input SIFT vectors.

Taking this heavily biased observation into account, we present an empirical estimation procedure based on the existing literature, which is both simple and practical.

\subsection{Indicative value}
The literature on multi-table hashing~\cite{tpami_norouzi2014, focs_greene1994} suggests that an indicative number,
$B/\log_2 N$, can be used to divide the table.
PQTable differs from previous studies as we are using PQ codes; however this indicative number can provide a good empirical estimate of the optimal $T$.
Because $T$ is a power of two in the proposed table, we quantize the indicative number into a power of two, and the final optimal $T^*$ is given as:
\begin{equation}
T^* = 2^{Q \left ( \log_2 ( B / \log_2 N) \right ) },
\label{eq:opt_t}
\end{equation}
where $Q(\cdot)$ is the rounding operation.
A comparison with the observed optimal number is shown in \Tref{tbl:table_div}.
In many cases, the estimated $T^*$ is a good estimation of the actual number, and the error margin was small even if the estimation failed,
as in the case of $B=32$ and $N=10^6$ (see \Fref{fig:anal1}).
Selected $T^*$ values are plotted as gray dots in \Fref{fig:anal1} and \ref{fig:anal2}.

\begin{table}
	\begin{center}
		\caption{Estimated and the actually observed best $T$ for SIFT1B data.}
		\begin{tabular}{@{}llllllllll@{}} \toprule
			& &  \multicolumn{8}{c}{N} \\ \cmidrule(l){3-10}
			$B$ & How & $10^2$ & $10^3$ & $10^4$ & $10^5$ & $10^6$ & $10^7$ & $10^8$ & $10^9$ \\ \midrule
			\multirow{2}{*}{32} & Observed & 4 & 4 & 2 & 2 & 1 & 1 & 1 & 1 \\
			& $T^*$ & 4 & 4 & 2 & 2 & 2 & 1 & 1 & 1 \\ \midrule
			\multirow{2}{*}{64}& Observed & 8 & 8 & 8 & 4 & 4 & 4 & 2 & 2 \\
			& $T^*$ & 8 & 8 & 4 & 4 & 4 & 2 & 2 & 2 \\  \bottomrule
		\end{tabular}
		\label{tbl:table_div}
	\end{center}
\end{table}

\section{Experimental Results}
\label{sec:exp}

In this section, we present our experimental results.
After the settings of the experiments are presented (\Sref{sec:exp_setting}), 
we evaluate several aspects of our proposed PQTable, including the analysis of runtime (\Sref{sec:exp_runtime}), accuracy (\Sref{sec:exp_accuracy}), and memory (\Sref{sec:exp_memory}).
Finally, the relationship between the proposed PQTable and dimensionality reduction is discussed (\Sref{sec:exp_distri}).

\begin{figure*}
	\begin{center}
		\subfloat[SIFT1B, $B=32$]{\includegraphics[width=0.45\linewidth]{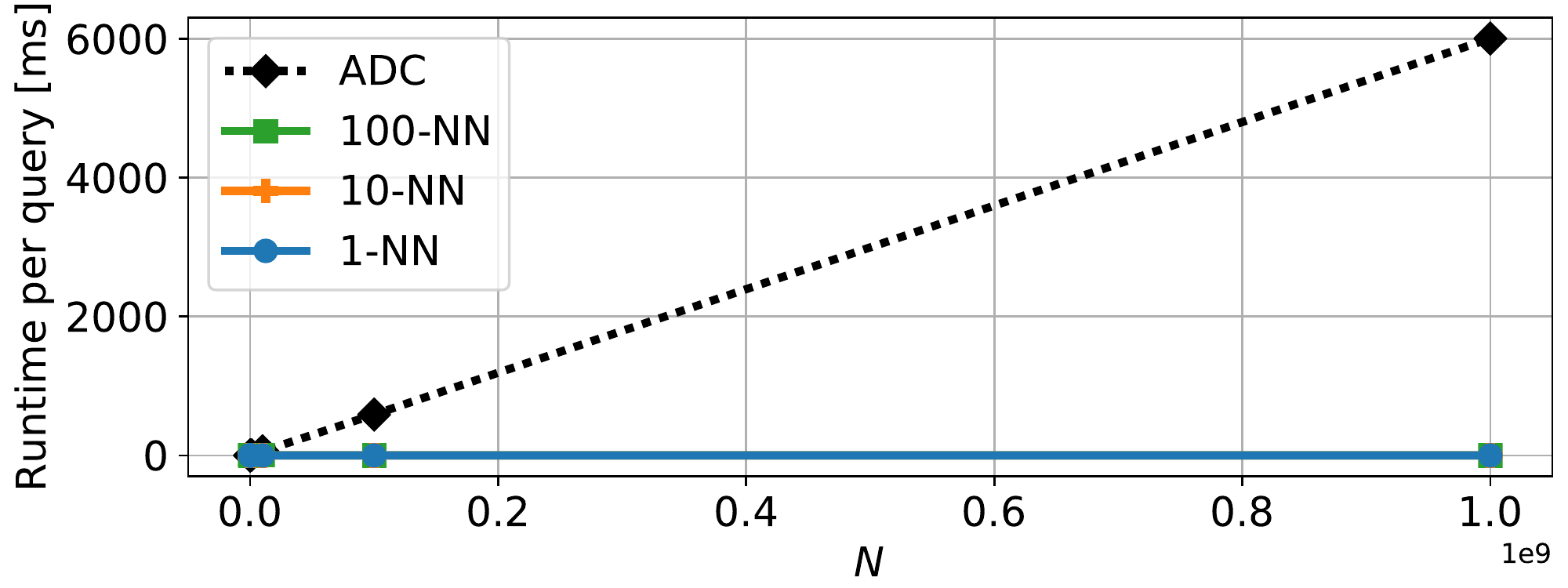}
			\label{fig:n_time_linear_sift1b_pq_b32}}
		\qquad
		\subfloat[SIFT1B, $B=64$]{\includegraphics[width=0.45\linewidth]{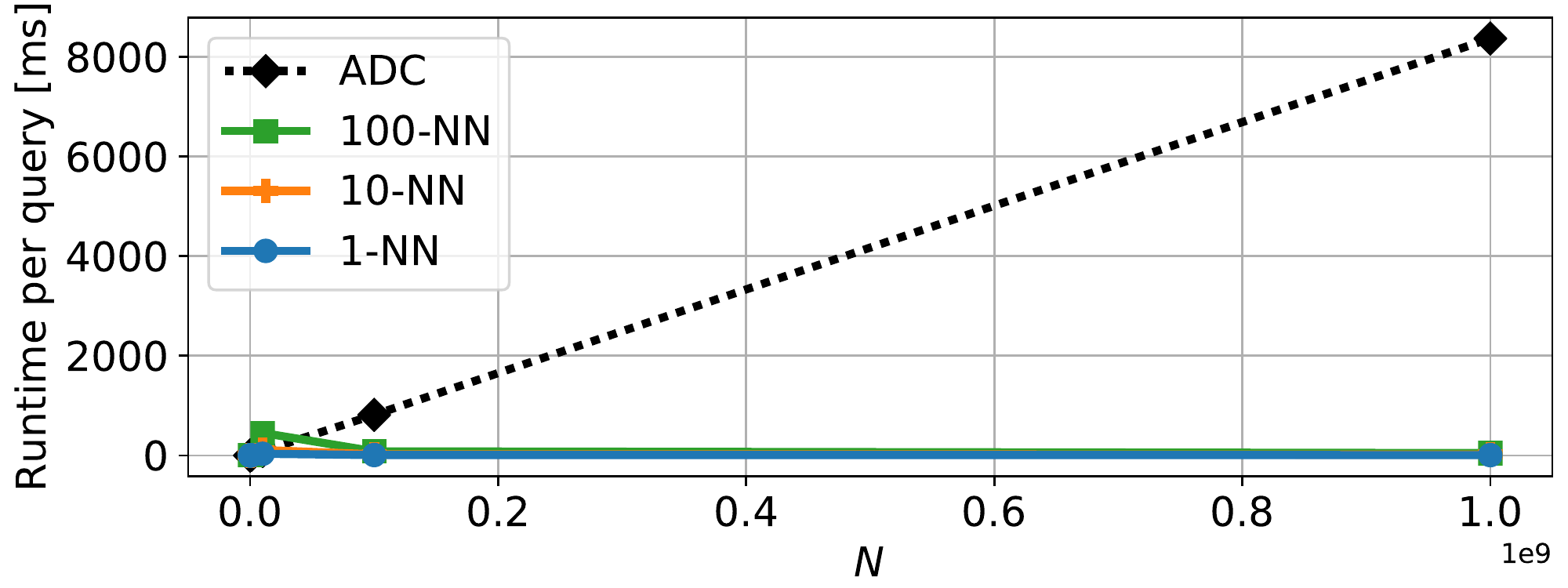}
			\label{fig:n_time_linear_sift1b_pq_b64}}
		
		\subfloat[SIFT1B, $B=32$, Log--log plot]{\includegraphics[width=0.45\linewidth]{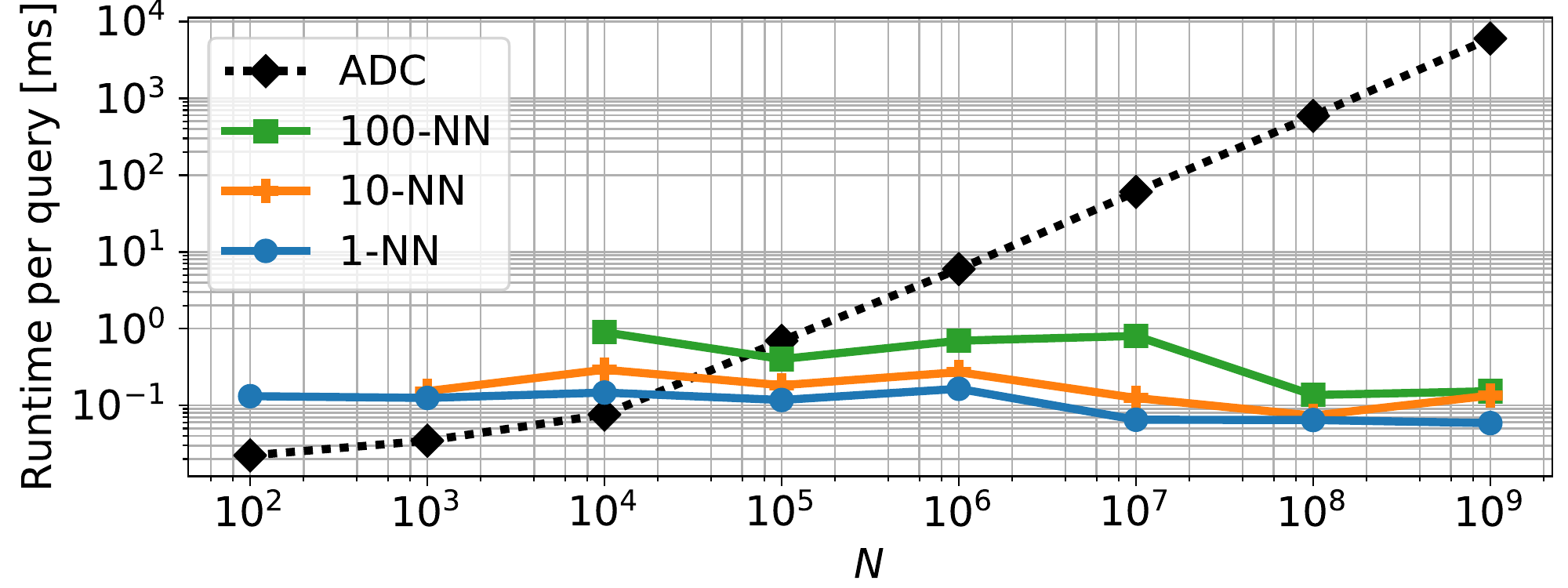}
			\label{fig:n_time_sift1b_pq_b32}}
		\qquad
		\subfloat[SIFT1B, $B=64$, Log--log plot]{\includegraphics[width=0.45\linewidth]{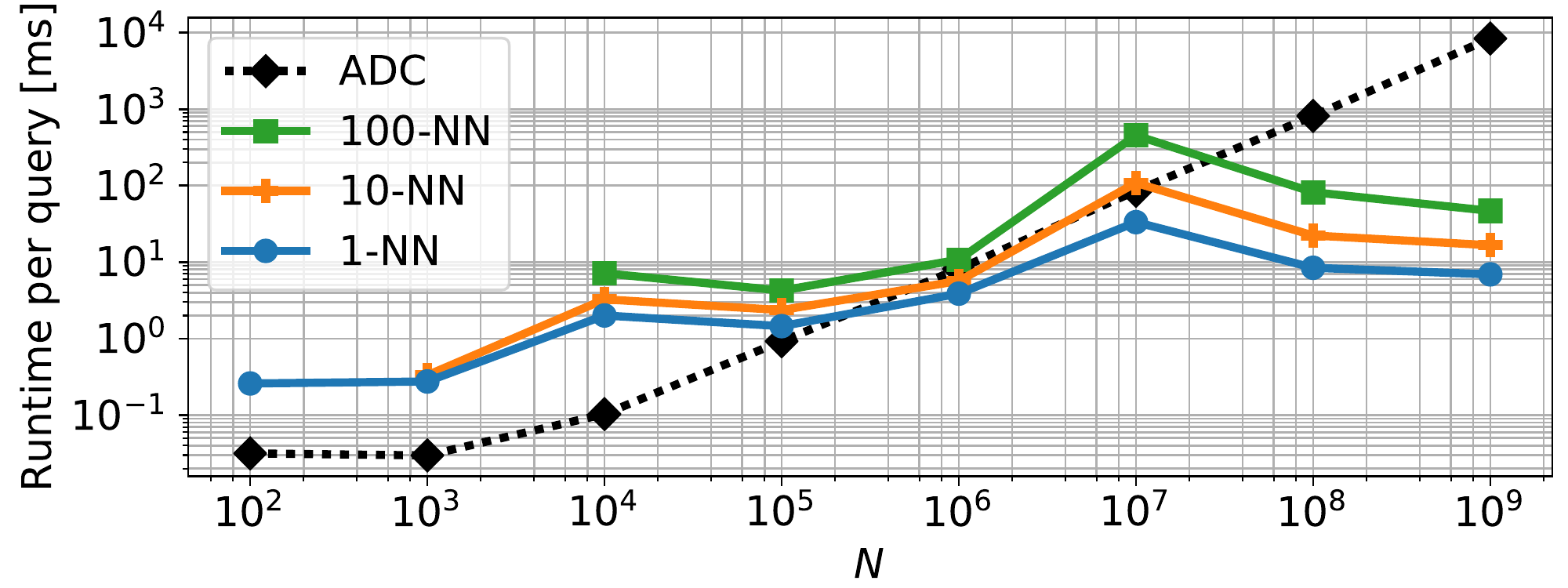}
			\label{fig:n_time_sift1b_pq_b64}}
		
		\subfloat[Deep1B, $B=32$, Log--log plot]{\includegraphics[width=0.45\linewidth]{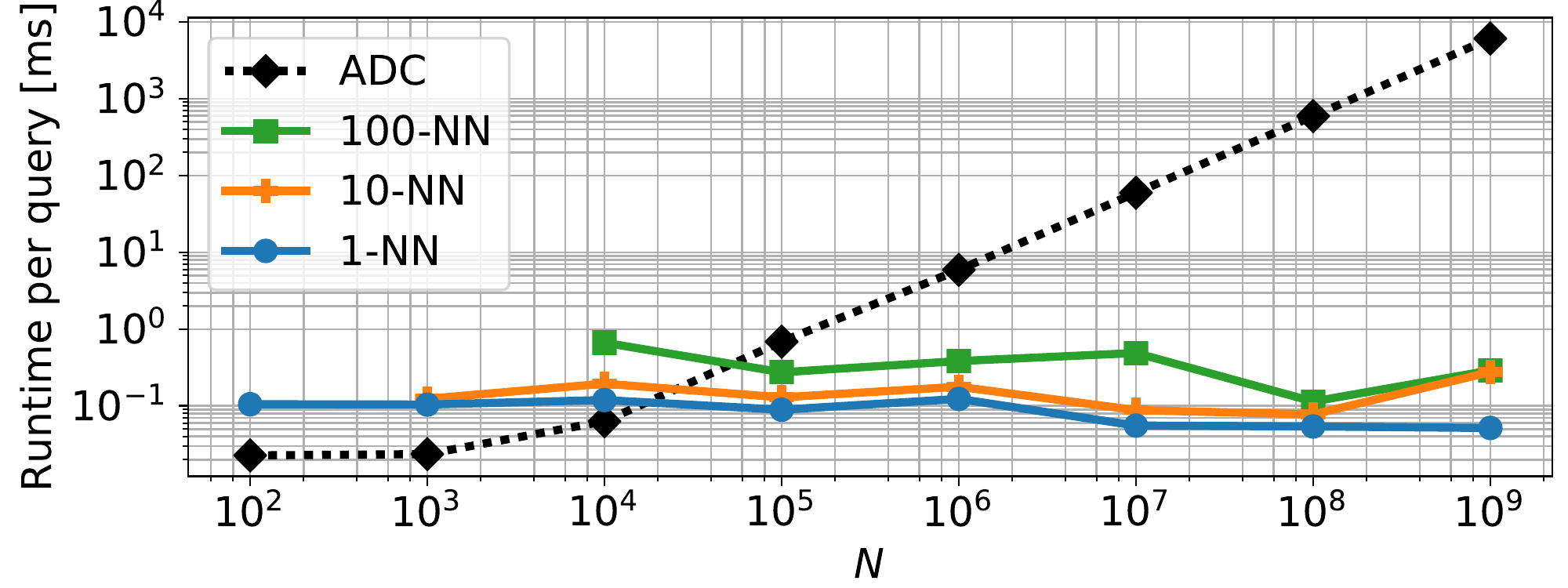}
			\label{fig:n_time_deep1b_pq_b32}}				
		\qquad
		\subfloat[Deep1B, $B=64$, Log--log plot]{\includegraphics[width=0.45\linewidth]{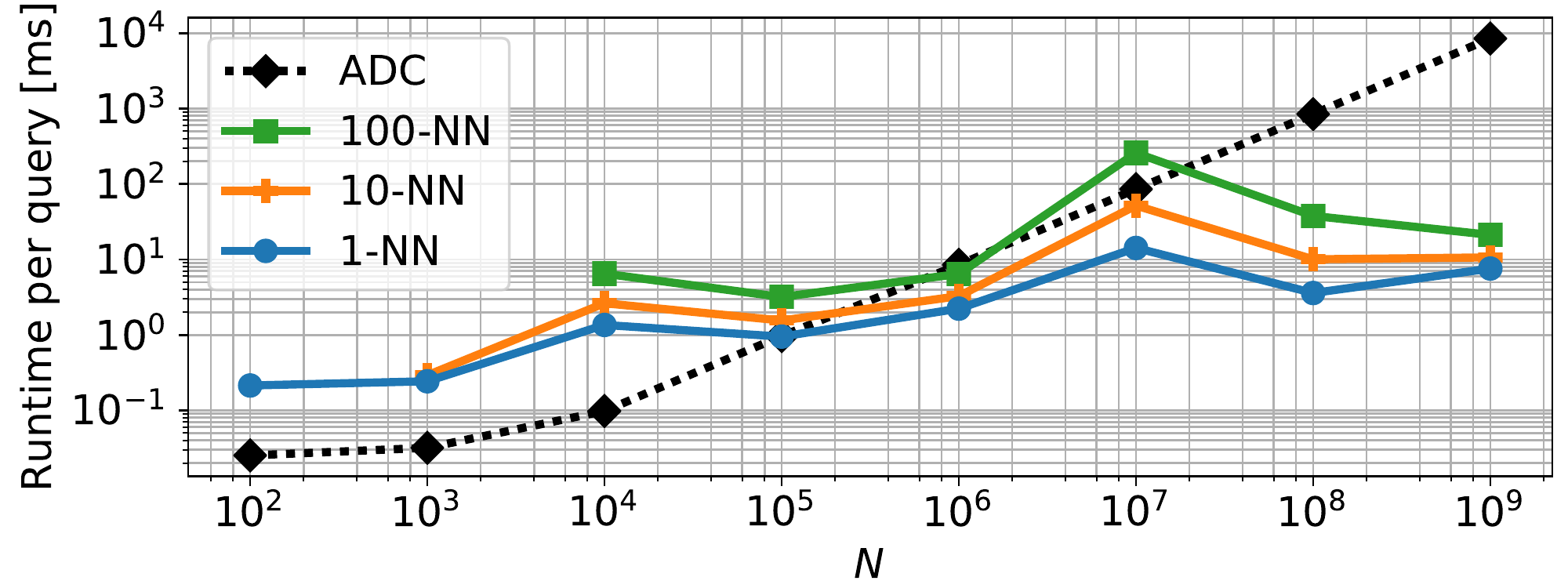}
			\label{fig:n_time_deep1b_pq_b64}}
		
	\end{center}
	\caption{Runtimes per query for the proposed PQTable with 1-, 10-, and 100-NN, and a linear ADC scan.}
	\label{fig:n_time_pq}
\end{figure*}

\subsection{Settings}
\label{sec:exp_setting}
We evaluated our approach using three datasets, SIFT1M, SIFT1B, and Deep1B.
All reported scores are values averaged over a query set.

SIFT1M dataset consists of 10K query, 100K training, and 1M base features.
Each feature is a 128D SIFT vector, where each element has a value ranging between 0 and 255.
The codewords $C$ are learned using the training data.
The base data are PQ-encoded and stored as a PQTable in advance.
SIFT1B is also a dataset of SIFT vectors, including
10K query, 100M training, and 1B base vectors.
Note that the top 10M vectors from the training features are used for learning $C$.
SIFT1M and SIFT1B datasets are from BIGANN datasets~\cite{icassp_jegou2011}.

The Deep1B dataset~\cite{cvpr_babenko2016} contains 10K query, 350M training, and 1B base features.
Each feature was extracted from the last fully connected layer of
GoogLeNet~\cite{cvpr_szegedy2015} for one billion images. 
The features were compressed by PCA to 96 dimensions and $l_2$ normalized. Each element in a feature can be a negative value.
For training, we used the top 10M vectors. 

In all experiments, $T$ is automatically determined by \Eref{eq:opt_t}.
All experiments were performed on a server with 3.6 GHz Intel Xeon CPU (6 cores, 12 threads) and 128 GB of RAM.
For training $C$, we use a multi-thread implementation.
To run the search, all methods are implemented with a single-thread for a fair comparison.
All source codes are available on \url{https://github.com/matsui528}.

\subsection{Runtime analysis}
\label{sec:exp_runtime}

\Fref{fig:n_time_pq} shows the runtimes per query for the proposed PQTable and a linear ADC scan. 
The results for SIFT1B with $B=32$ codes are presented in
\Fref{fig:n_time_linear_sift1b_pq_b32} (linear plot) and \Fref{fig:n_time_sift1b_pq_b32} (log-log plot).
The runtime of ADC depends linearly on $N$ and was fast for a small $N$, but required 6.0 s to scan $N=10^9$ vectors.
Alternatively, the PQTable ran for less than 1ms in all cases.
Specifically, the result of $N=10^9$ with 1-NN was $10^5$ times faster than that of the ADC.
The runtime and the speed-up factors against ADC are summarized in \Tref{tbl:speedup}.

The results with $B=64$ codes are presented in \Fref{fig:n_time_linear_sift1b_pq_b64} and \Fref{fig:n_time_sift1b_pq_b64}.
The speed-up over ADC was less dramatic than that of $B=32$, but was still $10^2$ to $10^3$ times 
faster when $10^8 < N$, which is highlighted in the linear plot (\Fref{fig:n_time_linear_sift1b_pq_b64}).

\Fref{fig:n_time_deep1b_pq_b32} and \Fref{fig:n_time_deep1b_pq_b64} illustrate the results for Deep1B dataset
with $B=32$ and $B=64$, respectively.
Notably, the runtimes for Deep1B show a similar tendency as those for SIFT1B, even though the distribution of SIFT features and GoogLeNet features are completely different.

\subsection{Accuracy}
\label{sec:exp_accuracy}
\begin{table*}
	\begin{center}
		\caption{The runtime performance of PQTable with $N=10^9$. The accuracy, runtime, and speed-up factors against ADC are presented.}
		\begin{tabular}{@{}llllllllll@{}} \toprule
			& & \multicolumn{3}{c}{Recall} & \multicolumn{4}{c}{Runtime / \textbf{Speed-up factors vs. ADC}} & \\ \cmidrule(lr){3-5} \cmidrule(lr){6-9}
			Data & $B$ & @1 & @10 & @100 & ADC & 1-NN & 10-NN & 100-NN & Memory [GB]\\ \midrule
			\multirow{2}{*}{SIFT1B} & 32 & 0.002 & 0.016 & 0.080 & 6.0 s / $\mathbf{1.0}$ & 0.059 ms / $\mathbf{1.0\times10^5}$ & 0.13 ms / $\mathbf{4.6\times10^4}$ & 0.15 ms / $\mathbf{4.0\times10^4}$ & 5.5 \\ 
			& 64 & 0.059 & 0.237 & 0.571 & 8.4 s / $\mathbf{1.0}$  & 6.9 ms / $\mathbf{1.2\times10^3}$ & 16.6 ms / $\mathbf{5.1\times10^2}$ & 46.8 ms / $\mathbf{1.8\times10^2}$ & 19.8 \\ \midrule
			\multirow{2}{*}{Deep1B} & 32 & 0.004 & 0.022 & 0.065 & 6.5 s / $\mathbf{1.0}$  & 0.081 ms / $\mathbf{8.0\times10^4}$ & 0.15 ms / $\mathbf{4.3\times10^4}$ & 0.16 ms / $\mathbf{4.1\times10^4}$ & 5.4 \\ 
			& 64 & 0.079 & 0.186 & 0.338 & 9.3 s / $\mathbf{1.0}$  & 8.7 ms / $\mathbf{1.1\times10^3}$ & 19.1 ms / $\mathbf{4.9\times10^2}$ & 54.3 ms / $\mathbf{1.7\times10^2}$ & 20.9 \\ \bottomrule 
		\end{tabular}
		\label{tbl:speedup}
	\end{center}
\end{table*}

\Tref{tbl:speedup} illustrates the accuracy (Recall@1) of PQTable and ADC for $N=10^9$.
In all cases, the accuracy of PQTable is as the same as that of ADC.

As expected, the recall@1 of $B=32$ is low (0.002 for SIFT1B, $B=32$) because a vector is highly compressed into a 32 bit PQ-code.
However, the search of over one billion data points was finished within just 0.059 ms.
These remarkably efficient results suggest that PQTable is one of
the fastest search schemes to date for billion-scale datasets on a single CPU.
Furthermore, because the data points are highly compressed, the required memory usage is just 5.5 GB.
Such fast searches with highly compressed data would be useful in cases where the runtime and the memory consumption take precedence over accuracy.

Note that the results with $B=64$ are comparable to the state-of-the-art
inverted-indexing-based methods;
e.g., 0.571 for PQTable v.s. 0.776 for OMulti-D-OADC-Local~\cite{tpami_babenko2015} (recall@100),
even though the PQTable does not require any parameter tunings.

\subsection{Memory consumption}
\label{sec:exp_memory}
\begin{figure}
	\begin{center}
		\includegraphics[width=1.0\linewidth]{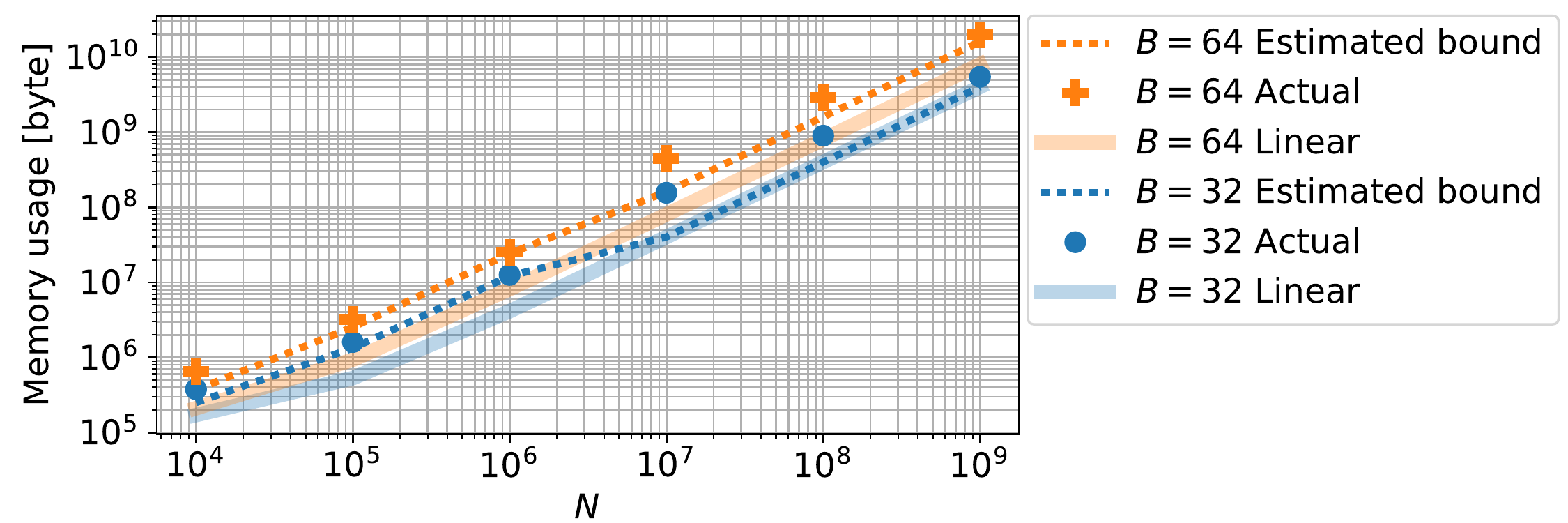}
	\end{center}
	\caption{Memory usage for the tables using the SIFT1B data. The dashed lines represent the theoretically estimated lower bounds. The circles and crosses represent the actual memory consumption for 32 and 64-bit tables. In addition, the linearly stored case for the ADC scan is shown.}
	\label{fig:memory}
\end{figure}
We show the estimated and actual memory usage of the PQTable in \Fref{fig:memory}.
Concrete values for $N=10^9$ are presented in \Tref{tbl:speedup}.
For the case of a single table ($T=1$), the theoretical memory usage involves the identifiers (4 bytes for each) in the table and the centroids of the PQ codes.
For multi-table cases, each table needs to hold the identifiers,
and the PQ codes themselves.
Using \Eref{eq:opt_t}, this theoretical lower-bound memory consumption (bytes) is summarized as: 
\begin{equation}
\begin{cases}
4N + 4DK & \mathrm{if}~~T^* = 1. \\
(4T^* + \frac{B}{8})N + 4DK & \mathrm{else}.
\end{cases}
\end{equation}
As a reference, we also show the cases where codes are linearly stored for a linear ADC scan
($BN/8+4KD$ bytes) in \Fref{fig:memory}.

For the $N=10^9$ with $B=64$ case, the theoretical memory usage is 16 GB,
and the actual cost is 19.8 GB. This difference comes from an overhead for the data structure of hash tables.
For example, 32-bit codes in a single table directly holding $2^{32}$ entries in an array require 32 GB of memory, even if all elements are empty.
This is due to a NULL pointer requiring 8 bytes with a 64-bit machine. 
To achieve more efficient data representation, we employed a sparse direct-address table~\cite{tpami_norouzi2014} as the data structure,
which enabled the storage of $10^9$ data points with a small overhead
and provided a worst-case runtime of $O(1)$. 

When PQ codes are linearly stored,
only 8 GB for $N=10^9$ with $B=64$ are required.
Therefore, we can say there is a trade-off among the proposed PQTable and
the linear ADC scan in terms of runtime and memory footprint (8.4 s with 8 GB v.s. 6.9 ms with 19.8 GB).

\subsection{Distribution of each component of the vectors}
\label{sec:exp_distri}
\begin{figure}
	\begin{center}
		\subfloat[Original SIFT data. Runtime per query: 2.9 ms. Recall@1=0.224 ]{\includegraphics[width=1.0\linewidth]{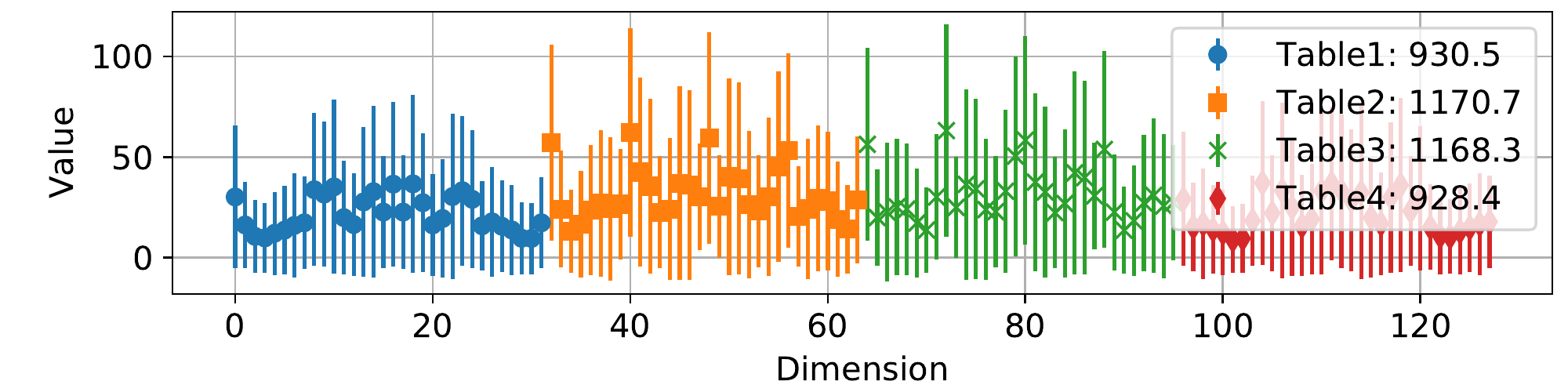}
			\label{fig:pca1}}
		
		\subfloat[PCA-aligned SIFT data. Runtime per query: 6.1 ms. Recall@1=0.117]{\includegraphics[width=1.0\linewidth]{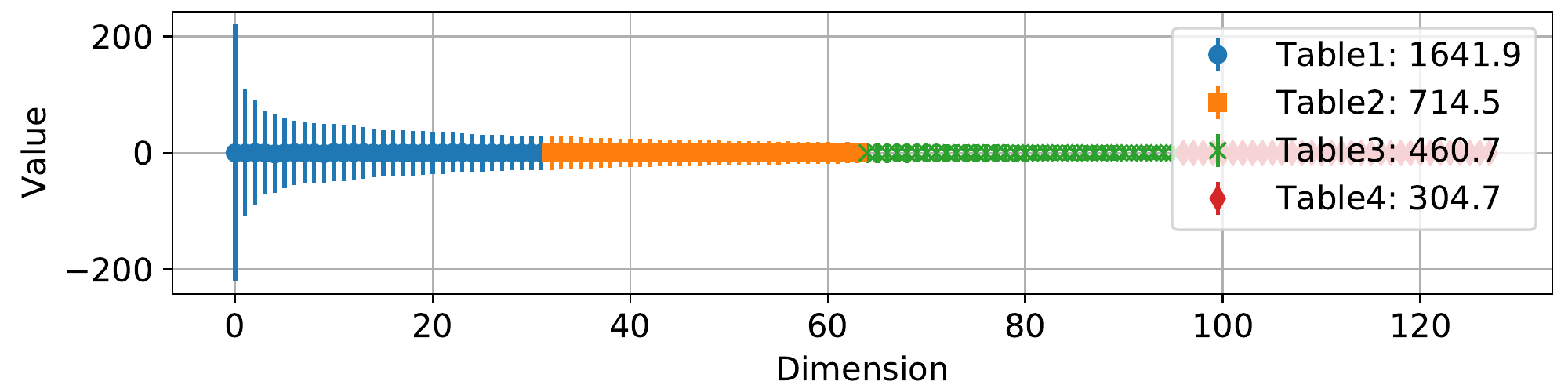}
			\label{fig:pca2}}
	\end{center}
	\caption{The average and standard deviation for the original SIFT vectors and the PCA-aligned vectors.}
	\label{fig:pca}
\end{figure}

Finally, we investigated how the distribution of vector components affects search performance, particularly for the multi-table case.
We prepared the SIFT1M data for the evaluation.
Principal-component analysis (PCA) is applied the data to ensure the same number of dimensionality (128).
\Fref{fig:pca1} shows the average and standard deviation for each dimension of the original
SIFT data, and \Fref{fig:pca2} presents that of the PCA-aligned SIFT data.
In both cases, a PQTable with $T=4$ and $B=64$ was constructed.
The dimensions associated with each table were plotted using the same color, and the sum of the standard deviations for each table is shown in the legends.

As shown in \Fref{fig:pca1}, the values for each dimension are distributed almost equally, which is the best case scenario for our PQTable.
Alternatively, \Fref{fig:pca2} shows a heavily biased distribution,
which is not desirable as the elements in Tables 2, 3, and 4 have almost no meaning.
In such a situation, however, the search is only two times slower than for the original SIFT data (2.9 ms for the original SIFT v.s. 6.1 ms for the PCA-aligned SIFT).
From this, we can say the PQTable remains robust for heavily biased element distributions.
Note that the recall value is lower for the PCA-aligned case because
PQ is less effective for biased data~\cite{tpami_jegou2011}.

\section{Extension to OPQTable}
\label{sec:extension_opqtable}

In this section,
we incorporate Optimized Product Quantization (OPQ)~\cite{tpami_ge2014} into PQTable (\Sref{sec:exp_opqtable}).
We then show a comparison to existing methods (\Sref{sec:exp_comp}).

\subsection{Optimized Product Quantization Table}
\label{sec:exp_opqtable}

\begin{figure*}
	\begin{center}
		\subfloat[SIFT1B, $B=32$, Log--log plot]{\includegraphics[width=0.45\linewidth]{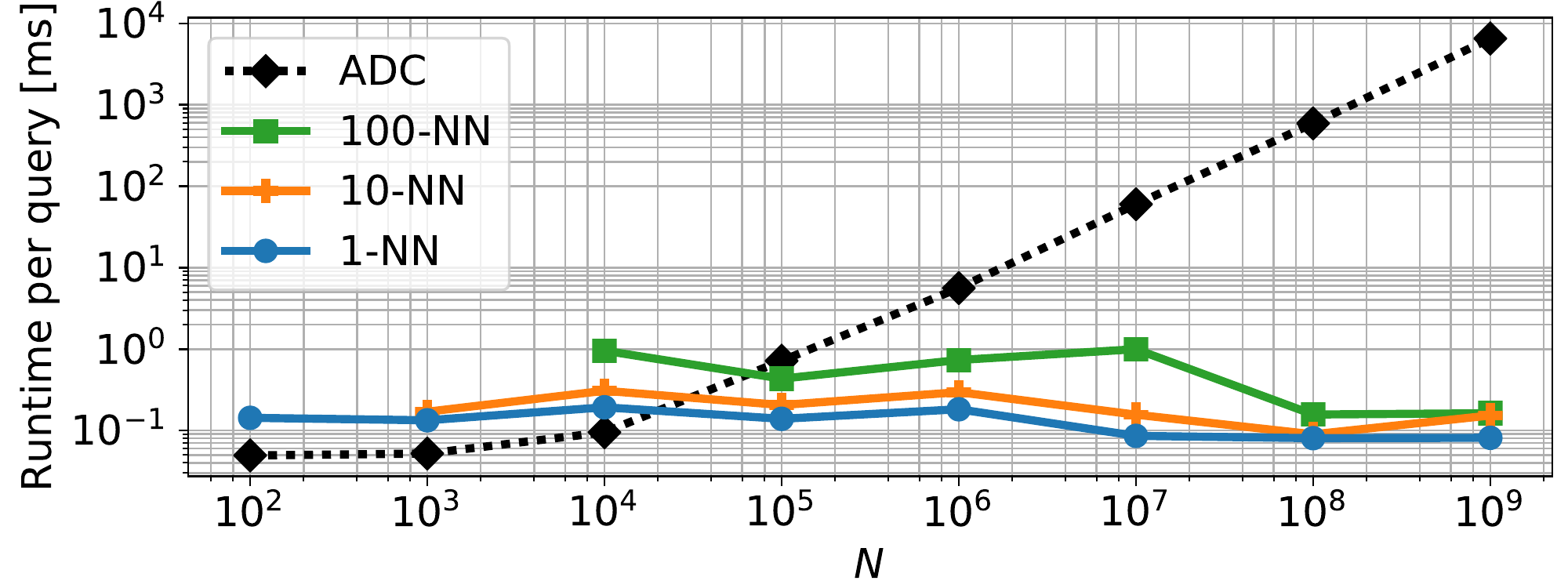}
			\label{fig:n_time_sift1b_opq_b32}}
		\qquad
		\subfloat[SIFT1B, $B=64$, Log--log plot]{\includegraphics[width=0.45\linewidth]{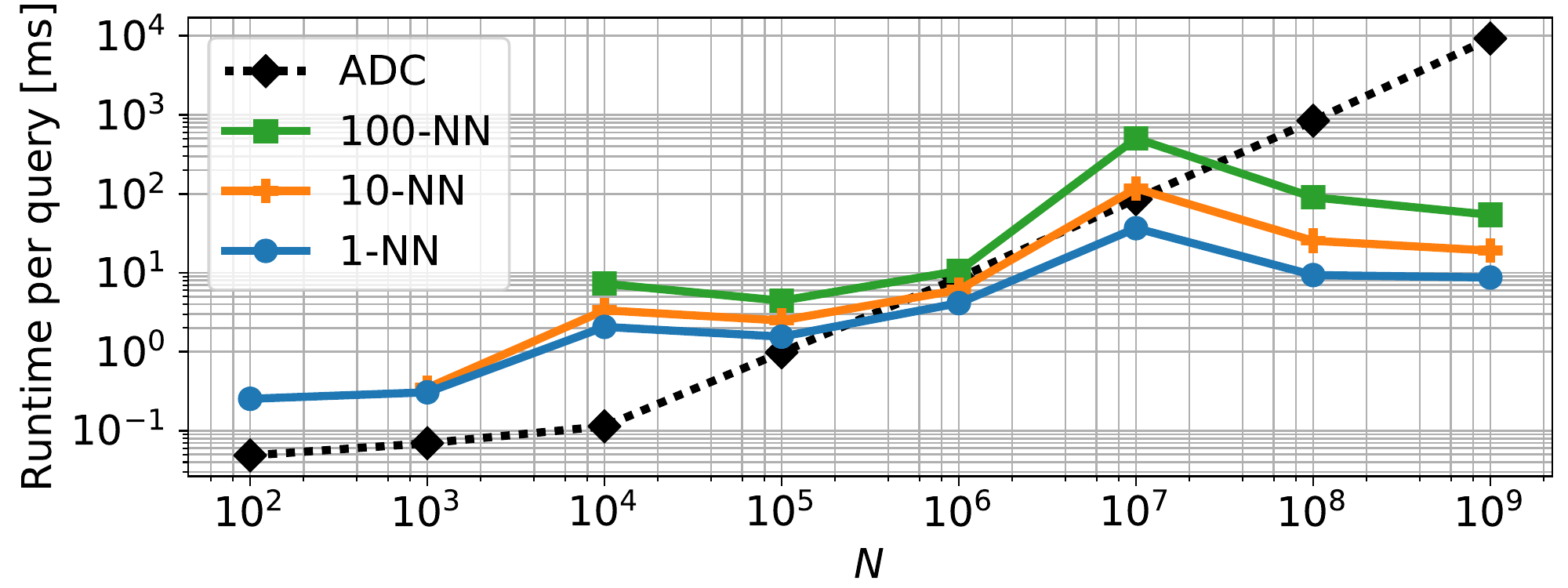}
			\label{fig:n_time_sift1b_opq_b64}}
		
	\end{center}
	\caption{Runtimes per query for the proposed OPQTable with 1-, 10-, and 100-NN, and a linear ADC scan.}
	\label{fig:n_time_opq}
\end{figure*}

OPQ~\cite{cvpr_norouzi2013, tpami_ge2014} is a simple yet effective extension of PQ.
In the offline encoding phase, database vectors are pre-processed by applying a rotation(orthogonal) matrix, and the rotated vectors are then simply PQ encoded.
In the online search phase, the rotation matrix is applied to a query vector, and the search is then performed in the same manner as PQ.

Our PQTable framework can naturally handle OPQ codes (we call this an OPQTable).
\Fref{fig:n_time_opq} illustrates the runtime performance of the OPQTable.
Compared to the PQTable, the OPQTable requires an additional $D \times D$ matrix multiplication for each query ($O(D^2)$).
In the SIFT1B and Deep1B datasets, this additional cost is not significant.
The runtimes are similar to that of the PQTable.
In addition, we found that accuracy is slightly but steadily better than that of the PQTable.
For SIFT1B, 0.002 (PQTable) v.s. 0.003 (OPQTable) with $B=32$, and 0.059 v.s. 0.063 with $B=64$.
For Deep1B, 0.004 v.s. 0.006 with $B=32$, and 0.079 v.s. 0.082 with $B=64$.
In terms of memory usage, OPQTable requires storing an additional $D \times D$ matrix, $4D^2$ bytes.
This additional cost is also negligible for SIFT1B and Deep1B.

\subsection{Comparison with existing methods}
\label{sec:exp_comp}

\begin{table*}
	\caption{A performance comparison for different methods using the SIFT1B data with 64-bit codes. The bracketed values are from \cite{tpami_babenko2015}.
		Note that the runtime of PQTable and IVFADC is for 1-NN case.}
	\label{tbl:comparison}
	\centering
	\scalebox{0.8}{
		\begin{tabular}{@{}llllllllll@{}} \toprule
			& \multicolumn{2}{c}{Params} & \multicolumn{3}{c}{Recall} & & & \multicolumn{2}{c}{Requirement} \\ \cmidrule(lr){2-3} \cmidrule(lr){4-6} \cmidrule(l){9-10}
			System & $\#cell$ & List-len & @1 & @10 & @100 & Runtime [ms] & Memory [GB] & Params & Additional training steps \\ \midrule
			OPQTable & - & - & 0.063 & 0.247 & 0.579 & 8.7 & 19.9 & None & None \\
			IVFADC~\cite{tpami_jegou2011} & $2^{13}$ & $8 \times 10^6$ & 0.115 {\footnotesize (0.112)} & 0.395 {\footnotesize (0.343)} & 0.763 {\footnotesize (0.728)} &  209 {\footnotesize (155)} & (12) & $\#cell$, list-len & Coarse quantizer \\
			OMulti-D-OADC-Local~\cite{tpami_babenko2015} & $2^{14} \times 2^{14}$ & $10^4$ & (0.268) & (0.644) & (0.776) & (6) & (15) & $\#cell$, list-len & Coarse quantizer, local codebook \\ \bottomrule		
		\end{tabular}
	}
\end{table*}

\Tref{tbl:comparison} shows a comparison with existing systems for the SIFT1B dataset with 64-bit codes.
We compared our OPQTable with two short-code-based inverted indexing systems: IVFADC~\cite{tpami_jegou2011} and OMulti-D-OADC-Local~\cite{tpami_babenko2015, corr_babenko2014, cvpr_kalantidis2014}.
IVFADC is the simplest system, and can be regarded as the baseline.
The simple k-means is used as the coarse quantizer.
OMulti-D-OADC-Local is a current state-of-the-art system.
The coarse quantizer involves PQ~\cite{cvpr_babenko2012},
the space for both the coarse quantizer and the short code is optimized~\cite{tpami_ge2014},
and the quantizers are per-cell-learned~\cite{corr_babenko2014, cvpr_kalantidis2014}.

The table shows that IVFADC performs with better accuracy;
however, it is usually slower than the OPQTable (8.7 ms v.s. 209 ms).
IVFADC requires two parameters to be tuned; the number of space partitions ($\#cell$) and the length of the list for re-ranking (or, equivalently, the search range $w$).
This value must be decided regardless of $L$ (the number of items to be returned).
As we first discussed in \Fref{fig:teaser_ivfadc}, the decision of these parameters is not trivial, though the proposed OPQ does not require any parameter tunings.

The best-performing system was OMulti-D-OADC-Local; it achieved better accuracy and memory usage than the OPQTable, though the computational cost for both was similar (8.7 ms v.s. 6 ms).
To fully make use of OMulti-D-OADC-Local,
one must tune two parameters manually (the $\#cell$ and the list-length); $\#cell$ is a critical parameter.
The reported value $2^{14}\times2^{14}$ is the optimal for $N=10^9$ data.
However, it is not clear that this parameter works well for other $N$.
In addition, several training steps are required
for learning the coarse quantizer, and for constructing local codebooks, both of which are time-consuming.

From the comparative study, we can say there are
advantages and disadvantages for both the inverted indexing systems
and our proposed PQTable:
\begin{itemize}
	\item \textbf{Static vs. dynamic database:} For a large static database
	where users have enough time and computational resources for tuning parameters and training quantizers, the previous inverted indexing systems should be used.
	Alternatively, if the database changes dynamically,
	the distribution of vectors may vary over time and parameters may need to be updated often.
	In such cases, the proposed PQTable would be the best choice.
	\item \textbf{Ease of use:} The inverted indexing systems produce good results but are difficult for a novice user to handle because they
	require several tuning and training steps. The proposed PQTable is deterministic, stable, conceptually simple, and much easier to use,
	as users do not need to decide on any parameters.
	This would be useful if users would like to use an ANN method
	simply as a tool for solving problems in another domain,
	such as fast SIFT matching for a large-scale 3D reconstruction~\cite{cvpr_cheng2014b}.
\end{itemize}

\section{Conclusion} 
\label{sec:conclusion}
In this study, we proposed the PQTable, a non-exhaustive search method for finding the nearest PQ codes without parameter tuning.
The PQTable is based on a multi-index hash table, and includes candidate code generation and the merging of multiple tables.
From our analysis, we showed that the required parameter value $T$ can be estimated in advance.
An experimental evaluation showed that the proposed PQTable could compute results $10^2$ to $10^5$ times faster than the ADC-scan method.

\textbf{Limitations:} PQTable is no longer efficient for $\ge$ 128-bit codes.
For example, SIFT1B with $T=4$ took approximately 3 s per query;
this is still faster than ADC, but slower than the state-of-the-art~\cite{tpami_babenko2015}.
This lag was caused by the inefficiency of the merging process for longer bit codes, and handling these longer codes should be an area of focus for future work.
It is important to note that $\le$ 128-bit codes are practical for many applications, such as the use of 80-bit codes for
image-retrieval systems~\cite{tmm_spyromitros2014}.
Another limitation is the heavy memory usage of hash tabls, 
and constructing a memory efficient data structure should also be an area of focus for future work.

\appendices
\section{Key generator}
\label{sec:multisequence}

\begin{algorithm}[t]
	\SetKwProg{Fn}{Function}{}{End} 
	\SetKwProg{Mem}{Member}{}{End} 
	\Mem{}{
		$candidates \gets \emptyset$~~\codecomment{// Non-duplicate priority-queue (\Aref{alg:nonduplicate_pqueue})} \\
		$dmat \gets$ Empty 2D-array~~\codecomment{// 2D-array of tuple$(k, dist, pos)$.} \\ 
		$C = C^1\times \dots \times C^M \gets \{\bvec{c}_1^1, \dots, \bvec{c}_K^1 \} \times \dots \times \{\bvec{c}_1^M, \dots, \bvec{c}_K^M \},~~\bvec{c}_k^m \in \mathbb{R}^{D/M}$~~~\codecomment{// Codewords}
	}
	\Fn{$\mathrm{Init}$}{
		\KwIn{$\bvec{q} = [ (\bvec{q}^1)^\top, \dots, (\bvec{q}^M)^\top ]^\top \in \mathbb{R}^D$. 
		}
		\For{$m \gets 1~\mathrm{to}~M$}{
			\For{$k \gets 1~\mathrm{to}~K$}{
				$dmat[m][k] \gets$ tuple$(k, d(\bvec{q}^m, \bvec{c}^m_k)^2, \mathrm{nil})$ ~~ \codecomment{//{\small ~$k \gets k,~dist \gets d(\bvec{q}^m, \bvec{c}^m_k)^2,~pos\gets nil$}}
			}
			SortByDist$(dmat[m][:])$ \codecomment{//~Sort~$dmat[m][1], \dots, dmat[m][K]$, using $dmat[m][k].dist$ as a key} \\
			\For{$k \gets 1~\mathrm{to}~K$}{
				$dmat[m][k].pos \gets k$
			}
		}
		\codecomment{// Collect the first tuple for each $dmat[m]$, then create a vector} \\
		$\bvec{e} \gets [~dmat[1][1], dmat[2][1], \dots, dmat[M][1]~]$ \\
		$candidates.$Push$(\bvec{e})$ \\	
	}	
	\Fn{$\mathrm{NextKey}$}{
		\KwOut{$\bvec{\bar{x}}\in\{1, \dots, K\}^M,$~~\codecomment{// PQ-code} \\
			~~~~~~~~~~~$d\in\mathbb{R}$.~~\codecomment{// $d_{AD}(\bvec{q}, \bvec{x})$}}
		
		$\bvec{e} \gets candidates.$Pop$()$ \\
		\For{$m \gets 1~\mathrm{to}~M$}{
			$\bvec{e}_{next} \gets \bvec{e}$\\
			\codecomment{// {\small Update $e_{next}[m]$ in $\bvec{e}_{next}$ by fetching the next-nearest tuple from $dtable$} } \\
			$e_{next}[m] \gets dtable[m][~e[m].pos + 1~]$ \\
			$candidates.$Push$(\bvec{e}_{next})$
		}
		\KwRet{$\bvec{\bar{x}} = \left [ e[1].k, \dots, e[M].k \right ]^\top,~~d=\sum_{m=1}^M e[m].dist$}
	}
	\caption{Key generator.}
	\label{alg:msa}
\end{algorithm}

\begin{algorithm}[t]
	\SetKwProg{Fn}{Function}{}{End} 
	\SetKwProg{Mem}{Member}{}{End} 
	
	\Mem{}{
		$pqueue \gets \emptyset$~~\codecomment{// Priority-queue} \\
		$Z \gets \emptyset$~~\codecomment{// A set of integers}
	}
	
	\Fn{$\mathrm{Push}$}{
		\KwIn{$v\in \mathbb{R}, z\in \mathbb{N}$}
		\If{$z \notin Z$}{
			$Z \gets Z \cup z$ \\
			$pqueue$.Push($v$)
		}
	}
	
	\Fn{$\mathrm{Pop}$}{
		\KwOut{$v\in \mathbb{R}$}
		\codecomment{// The minimum $v$ in $pqueue$ is popped} \\
		\KwRet{$pqueue.\mathrm{Pop()}$}
	}
	\caption{Non-duplicate priority-queue}
	\label{alg:nonduplicate_pqueue}
\end{algorithm}

\begin{figure*}
	\begin{center}
		\includegraphics[width=1.0\linewidth]{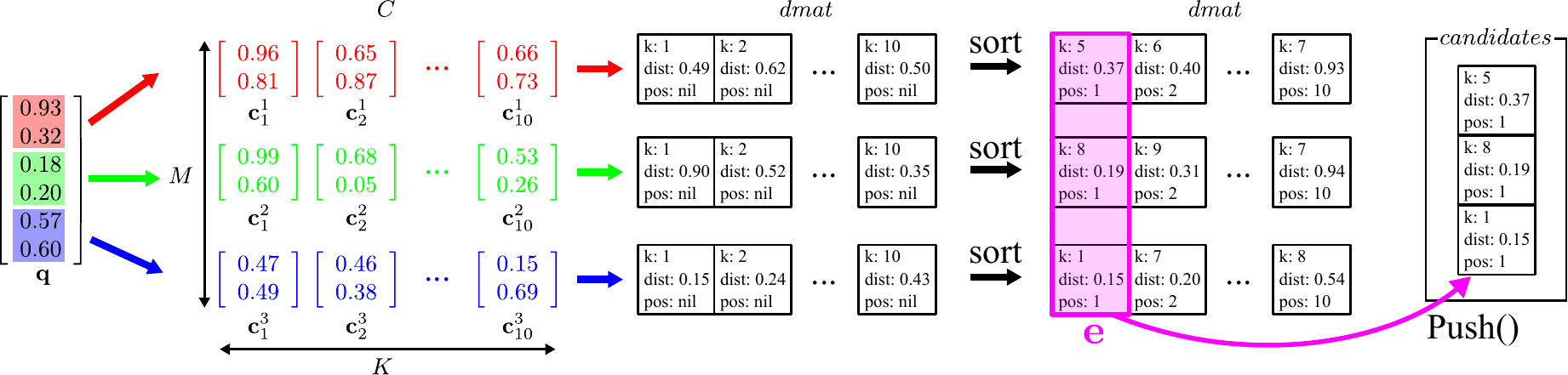}
	\end{center}
	\caption{An example of the initialization of the multisequence algorithm (\texttt{Init} in \Aref{alg:msa}), where $D=6$, $M=3$ and $K=10$.}
	\label{fig:msa1}
\end{figure*}

\begin{figure*}
	\begin{center}
		\subfloat[The first call.]{\includegraphics[width=0.3\linewidth]{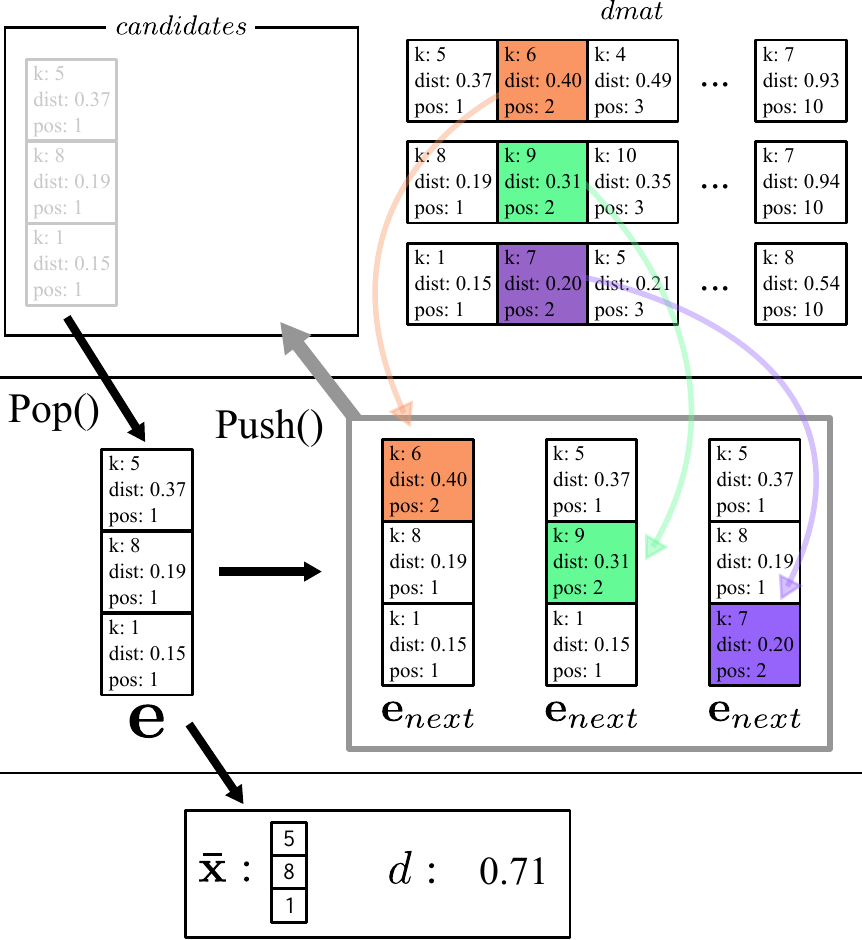}
			\label{fig:msa2_1}}
		~~~~~
		\subfloat[The second call.]{\includegraphics[width=0.3\linewidth]{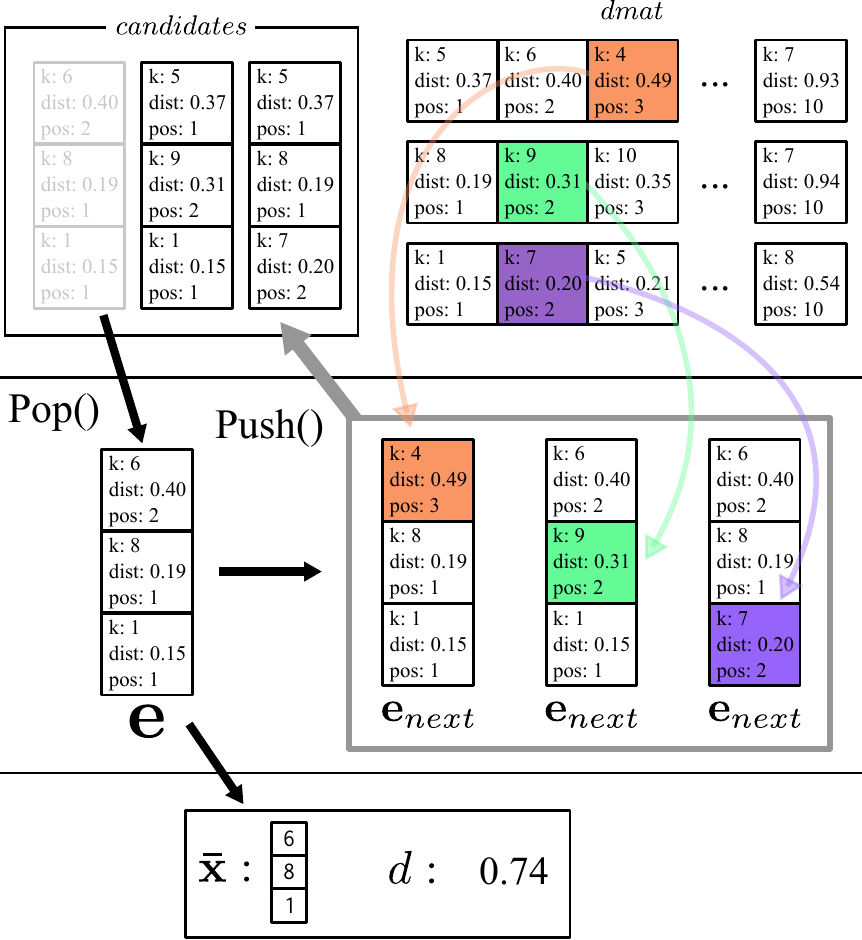}
			\label{fig:msa2_2}}
		~~~~~
		\subfloat[The third call.]{\includegraphics[width=0.3\linewidth]{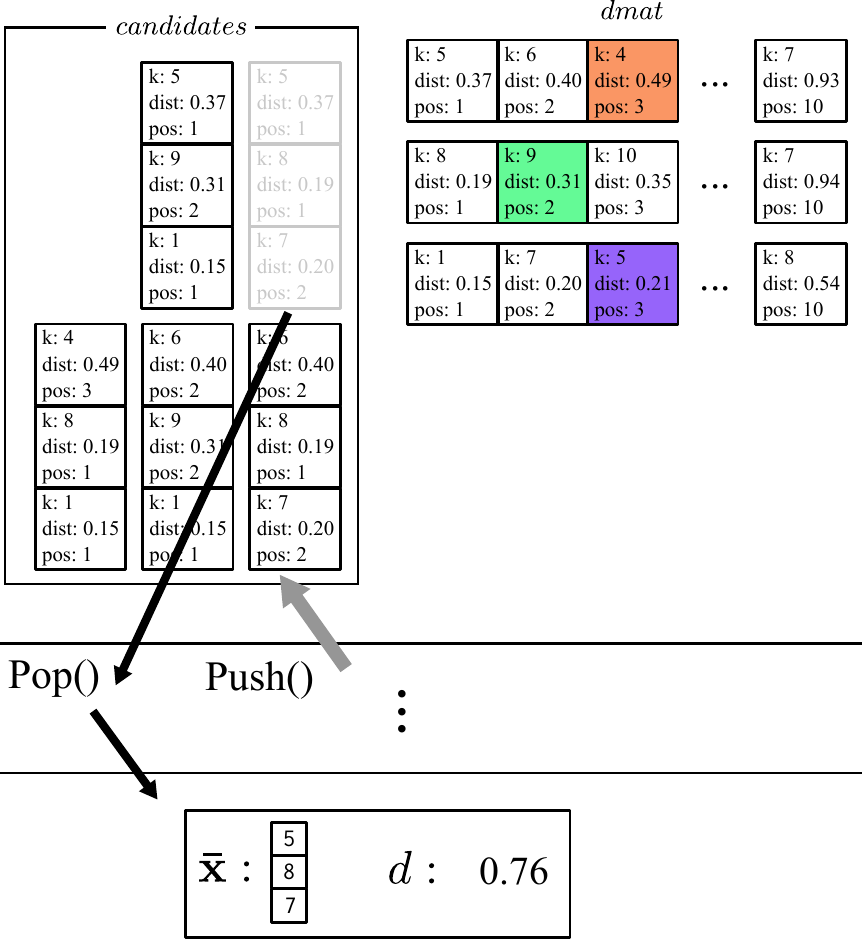}
			\label{fig:msa2_3}}
	\end{center}
	\caption{An example of the key generation step (\texttt{NextKey} in \Aref{alg:msa}).
		We assume that the initialization step is finished as shown in \Fref{fig:msa1}.
		The figures show the results when the \texttt{NextKey} function is called three times.
		\Fref{fig:msa2_1}, \Fref{fig:msa2_2}, and \Fref{fig:msa2_3} show the first, second, and third call, respectively.}
	\label{fig:msa2}
\end{figure*}

\Aref{alg:msa} introduces a data structure and an algorithm of the key generator.
This scheme is mathematically equivalent to the multi-sequence algorithm~\cite{cvpr_babenko2012}.
The data structure of the generator includes a non-duplicate priority-queued ($candidates$) and a 2D-array ($dmat$) (L2-L3). 
For a given query vector, the querying algorithm operates in two stages: initialization (\texttt{Init}, L5-L14) and key generation (\texttt{NextKey}, L15-L22).
Visual examples are shown in \Fref{fig:msa1} and \ref{fig:msa2}.

When the generator is instantiated, $candidates$, $dmat$, and $C$ are created (L1-L3).
$candidates$ is a priority-queue containing no duplicate items.
$dmat$ is an $M\times K$ 2D-array consisting of tuples, and each tuple consists of three scalars:$k\in \{1, \dots, K\}$, $dist \in \mathbb{R}$, and $pos \in \{1, \dots, K\}$; $C$
are codewords for PQ.
Note that $candidates$ and $dmat$ are created with empty elements.
$C$ is pre-trained and loaded.

\subsection{Initialization}
The initialization step takes a query vector $\bvec{q}\in \mathbb{R}^D$ as an input.
First, the $m$-th part of $\bvec{q}$ and $m$-th codewords $C^m$ are compared.
The resultant squared distances are recorded with $k$ in $dmat$ (L8), and  each row in $dmat$ is sorted by distance (L9).
After being sorted, the indices are recorded in $pos$ (L11).
Next, the first tuple from each row is picked to construct $\bvec{e}$ (L13). 
Finally, $\bvec{e}$ is inserted into $candidates$ (L14).
It is clear that the $k$s in $\bvec{e}$ can create a PQ-code. 
The entire computational cost of the initialization is $O(K(D + M\log K))$, which is negligible for a large database\footnote{If
	$D$ is sufficiently large, this cost is not negligible.
	Handling such case is a hot topic~\cite{cvpr_zhang2015}, but out of the scope of this paper.}
\Fref{fig:msa1} illustrates this process.

Note that $candidate$ is a priority queue without duplicate items.
We show a data structure and functions over the structure in \Aref{alg:nonduplicate_pqueue}.
This data structure holds a usual priority queue $pqueue$ and a set of integers $Z$.
We assume an item has two properties: $v\in\mathbb{R}$ and $z\in\mathbb{N}$.
As with a normal priority queue, $v$ shows the priority of an item, whereas $z$ is used as an identifier to distinguish one item from another.
When the \texttt{Push} function is called, $z$ of a new item is checked to determine whether it is $Z$ or not.
If $z$ already exists, the item is not inserted.
If $z$ is not present in $Z$, the item is inserted to $pqueue$, and $z$ is also inserted in $Z$.
The \texttt{Pop} function is as the same as a normal priority queue; the item with the minimum $v$ is popped and returned.
If we denote $Q$ as the number of items in $pqueue$, 
the \texttt{Push} takes $O(\log Q)$ on average and $O(\log Q + |Z|)$ in the worst case, using a hash table to represent $Z$.
The \texttt{Pop} takes $O(\log Q)$.

In \Aref{alg:msa}, we inserted not only $v$ and $z$, but also a vector of tuples ($\bvec{e}$ in L13.)
The sum of square distances, $d=\sum_{m=1}^M e[m].dist$, is used as a priority $v$.
As an identifier $z$ of an item, the combination of $k$s from each tuples ($e[1].k, \dots, e[M].k$) is leveraged. 
This ``duplicate checking'' step is a different implementation to the original multi-sequence algorithm~\cite{cvpr_babenko2012}; in the original algorithm, an $M$-dimensional array is required to check the duplicates, consuming much more memory for a large $M$ value.
The results of the two algorithms were identical.

\subsection{Key generation:}
As previously mentioned, our purpose is to enumerate candidates of hashing one by one.
The first candidate is an original PQ-code of $\bvec{q}$ itself.
The second candidate should be a possible code (in $\{1, \dots, K\}^M$) whose distance to the query is the second nearest, and the third candidate’s distance should be the third nearest, etc.

Enumeration is achieved by maintaining $candidates$.
Whenever \texttt{NextKey} is called, the item with the minimum $d$ is popped from $candidates$, which we denote as $\bvec{e}$ (L16).
If we recall $\bvec{e}$ consists of $M$ tuples, it is obvious that we have $M$ possibilities for next-nearest codes.
Given a current $\bvec{e}$, we can slightly update $m$-th tuple for each $m \in \{1, \dots, M\}$,
making $M$ $\bvec{e}_{next}$.
This update is achieved by fetching the next tuple in $dtable$
because each row in $dtable$ are sorted in the ascending order of $d$ (L18-L20). 
$M$ $\bvec{e}_{next}$ are then pushed into $candidates$ (L21).
Finally, a PQ-code $\bvec{\bar{x}}$ is created by picking each $e[m].k$ for each $m$.
The PQ-code and $d$ is then returned (L22).
This \texttt{NextKey} process is illustrated in \Fref{fig:msa2}.

\section{Proof that required identifiers are already marked}
\label{sec:proof}
We proved that all items whose asymmetric distance ($d_{AD}$) is less that $d_{min}$ must be marked
in the querying process of a multi-PQTable.
Hereinafter, we denote the $d_{AD}$ from the $t$th part as $d_{AD}^t(\bvec{q}, \bvec{x})$, which leads to:
\begin{equation}
\sum_{t=1}^T d_{AD}^t(\bvec{q}, \bvec{x})^2 = d_{AD}(\bvec{q}, \bvec{x})^2.
\end{equation}
Let us assume that the identifier $n^*$ such that $c[n^*]=T$ is found (L24 in \Aref{alg:multi_table}).
We define $d_{min}=d_{AD}(\bvec{q}, \bvec{x}_{n^*})$ (L25).
In addition, we denote $\mathcal{N}^t=\{n^t\}$ as a set of identifiers which are marked when $t$th table is focused.
For example, in \Fref{fig:overview2}, $n^*=456$, $\mathcal{N}^1=\{585, 2, 456\}$, and $\mathcal{N}^2=\{24, 456\}$.
It is obvious that $d_{AD}^t(\bvec{q}, \bvec{x}_n) \le d_{AD}^t(\bvec{q}, \bvec{x}_{n^*})$ for any $n \in \mathcal{N}^t$
because of its construction.
Similarly, it is also clear that $d_{AD}^t(\bvec{q}, \bvec{x}_{n^*}) < d_{AD}^t(\bvec{q}, \bvec{x}_n)$ for any $n \notin \mathcal{N}^t$.
Next, we introduce a proposition.

\textit{Proposition:}
Any items $n$ which satisfied $d_{AD}(\mathbf{q}, \mathbf{x}_n) < d_{min}$ must be already marked.

\textit{Proof:}
This proposition is proved by contradiction.
Suppose there is an identifier $\hat{n}$, where $d_{AD}(\bvec{q}, \bvec{x}_{\hat{n}}) < d_{min}$, and $\hat{n}$ has not been marked ($\hat{n} \notin \mathcal{N}^t$ for all $t$).
Because $\hat{n}$ is not marked, $d_{AD}^t(\bvec{q}, \bvec{x}_{n^*}) < d_{AD}^t(\bvec{q}, \bvec{x}_{\hat{n}})$ for all $t$, as stated above.
Summing up all $t$ leads to $d_{AD}(\bvec{q}, \bvec{x}_{n^*}) = d_{min} < d_{AD}(\bvec{q}, \bvec{x}_{\hat{n}})$.
This contradicts the premise. 

\ifCLASSOPTIONcaptionsoff
  \newpage
\fi



\bibliographystyle{IEEEtran}
\bibliography{egbib}
\end{document}